\documentclass{bmvc2k}

\title{Adversarial Learning for \\
	Semi-Supervised Semantic Segmentation}

\addauthor{Wei-Chih Hung}{whung8@ucmerced.edu}{1}
\addauthor{Yi-Hsuan Tsai}{ytsai@nec-labs.com}{2}
\addauthor{Yan-Ting Liou}{lyt@csie.ntu.edu.tw}{34}
\addauthor{Yen-Yu Lin}{yylin@citi.sinica.edu.tw}{4}
\addauthor{Ming-Hsuan Yang}{mhyang@ucmerced.edu}{15}

\addinstitution{
	University of California, Merced
}
\addinstitution{
	NEC Laboratories America
}
\addinstitution{
	National Taiwan University
}

\addinstitution{
	Academia Sinica, Taiwan
}
\addinstitution{
	Google Cloud
}

\runninghead{Hung,  Tsai, Liou, Lin, Yang}{ADVERSARIAL LEARNING FOR SEMI-SEGMENTATION}

\def\etal{\emph{et al}\bmvaOneDot}

\usepackage{amsfonts}       
\usepackage{nicefrac}       
\usepackage{microtype}      
\usepackage{graphicx}

\usepackage{booktabs}       
\usepackage{paralist}
\usepackage{multirow}
\usepackage[normalem]{ulem}

\begin{document}
	
	\maketitle
	
	\begin{abstract}
		
		We propose a method for semi-supervised semantic segmentation using an adversarial network.
		While most existing discriminators are trained to classify input images as real or fake on the image level, we design a discriminator in a fully convolutional manner to differentiate the predicted probability maps from the ground truth segmentation distribution with the consideration of the spatial resolution.
		We show that the proposed discriminator can be used to improve semantic segmentation accuracy by coupling the adversarial loss with the standard cross entropy loss of the proposed model. 
		In addition, the fully convolutional discriminator enables semi-supervised learning through discovering the trustworthy regions in predicted results of unlabeled images, thereby providing additional supervisory signals.
		In contrast to existing methods that utilize weakly-labeled images, our method leverages unlabeled images to enhance the segmentation model.
		Experimental results on the PASCAL VOC 2012 and Cityscapes datasets demonstrate the effectiveness of the proposed algorithm.
	
	\end{abstract}
	
	\section{Introduction}
	

	Semantic segmentation aims to assign a semantic label, e.g., person, dog, or road, to each pixel in images.
	This task is of essential importance to a wide range of applications, such as autonomous driving and image editing.
	Numerous methods have been proposed to tackle this task~\cite{fcn,crfasrnn,deepparsing,dilated,piecewise,hung2017scene}, and abundant benchmark datasets have been constructed~\cite{pascal, pascal_context, cityscapes, zhou2016semantic}
	with focus on different sets of scene/object categories as well as various real-world applications.
	However, this task remains challenging because of large object/scene appearance variations, occlusions, and lack of context understanding.
	Convolutional Neural Network (CNN) based methods, such as the Fully Convolutional Network (FCN)~\cite{fcn}, have recently achieved significant improvement on the task of semantic segmentation, and most state-of-the-art algorithms are based on FCN and additional modules.
	
	Although CNN-based approaches have achieved astonishing performance, they require an enormous amount of training data.
	Different from image classification and object detection, semantic segmentation requires accurate per-pixel annotations for each training image, which can cost considerable expense and time.
	To ease the effort of acquiring high-quality data, semi/weakly-supervised methods have been applied to the task of semantic segmentation.
	These methods often assume that there are additional annotations on the image level~\cite{pinheiro2015weakly, papandreou2015weakly, hong2015decoupled, qi2016augmented, pathak2015constrained}, box level~\cite{dai2015boxsup}, or point level~\cite{bearman2016s}.

	In this paper, we propose a semi-supervised semantic segmentation algorithm based on adversarial learning.
	The recent success of Generative Adversarial Networks (GANs) \cite{gan} 
	facilitate effective unsupervised and semi-supervised learning in numerous tasks. 
	A typical GAN consists of two sub-networks, i.e., generator and discriminator, in which these two sub-networks play a min-max game in the training process. The generator takes a sample vector and outputs a sample of the target data distribution, e.g., human faces,
	while the discriminator aims to differentiate generated samples from target ones.
	The generator is then trained to confuse the discriminator through back-propagation and therefore generates samples that are similar to those from the target distribution.
	In this paper, we apply a similar methodology and treat the segmentation network as the generator in a GAN framework.
	Different from the typical generators that are trained to generate images from noise vectors, our segmentation network outputs the probability maps of the semantic labels given an input image.
	Under this setting, we enforce the outputs of the segmentation network as close as possible to the ground truth label maps spatially.

	To this end, we adopt an adversarial learning scheme and propose a fully convolutional discriminator that learns to differentiate ground truth label maps from probability maps of segmentation predictions.
	Combined with the cross-entropy loss, our method uses an adversarial loss that encourages the segmentation network to produce predicted probability maps close to the ground truth label maps in a high-order structure.
	The idea is similar to the use of probabilistic graphical models such as Conditional Random Fields (CRFs) \cite{crfasrnn,deeplab,piecewise}, but without the extra post-processing module during the testing phase.
	In addition, the discriminator is not required during inference, and thus the proposed framework does not increase any computational load during testing.
	By employing the adversarial learning, we further exploit the proposed scheme under the semi-supervised setting.

	In this work, we combine two semi-supervised loss terms to leverage the unlabeled data.
	%
	%
	First, we utilize the confidence maps generated by our discriminator network as the supervisory signal to guide the cross-entropy loss in a self-taught manner.
	The confidence maps indicate which regions of the prediction distribution are close to the ground truth label distribution so that these predictions can be trusted and trained by the segmentation network via a masked cross-entropy loss.
	Second, we apply the adversarial loss on unlabeled data as adopted in the supervised setting, which encourages the model to predict segmentation outputs of unlabeled data close to the ground truth distributions.

	The contributions of this work are summarized as follows. First, we develop an adversarial framework that improves semantic segmentation accuracy without requiring additional computation loads during inference.
	Second, we propose a semi-supervised framework and show that the segmentation accuracy can be further improved by adding images without any annotations.
	Third, we facilitate the semi-supervised learning by leveraging the discriminator network response of unlabeled images to discover trustworthy regions that facilitate 
	the training process for segmentation.
	Experimental results on the PASCAL VOC 2012~\cite{pascal} and Cityscapes~\cite{cityscapes} datasets validate the effectiveness of the proposed adversarial framework for semi-supervised semantic segmentation.
	
	\section{Related Work}

	{\flushleft \bf Semantic segmentation.}
	Recent state-of-the-art methods for semantic segmentation are based on the advances of CNNs. As proposed in \cite{fcn}, one can transform a classification CNN, e.g., AlexNet~\cite{alexnet}, VGG~\cite{vgg}, or ResNet~\cite{resnet}, to a fully-convolutional network (FCN) for the semantic segmentation task. 
	However, it is usually expensive and difficult to label images with pixel-level annotations. 
	To reduce the heavy efforts of labeling segmentation ground truth, numerous weakly-supervised approaches are proposed in recent years.
	In the weakly-supervised setting, the segmentation network is not trained at the pixel level with the fully annotated ground truth. Instead, the network is trained with various weak-supervisory signals that can be obtained easily.
	Image-level labels are exploited as the supervisory signal in most recent approaches.
	The methods in \cite{pinheiro2015weakly} and \cite{pathak2014fully} use Multiple Instance Learning (MIL) to generate latent segmentation label maps for supervised training.
	On the other hand, Papandreou~\etal~\cite{papandreou2015weakly} use the image-level labels to penalize the prediction of non-existent object classes, while Qi~\etal~\cite{qi2016augmented} use object localization to refine the segmentation. 
	Hong~\etal~\cite{hong2015decoupled} leverage the labeled images to train a classification network as the feature extractor for deconvolution.
	In addition to image-level supervisions, the segmentation network can also be trained with bounding boxes~\cite{dai2015boxsup, khoreva_CVPR17}, point supervision~\cite{bearman2016s}, or web videos~\cite{hong2017weakly}.

	However, these weakly supervised approaches do not perform as well as the fully-supervised methods especially because it is difficult to infer the detailed boundary information 
	from weak-supervisory signals.
	Hence semi-supervised learning is also considered in some methods to enhance the prediction performance.
	In such settings, a set of fully-annotated data and weakly-labeled samples 
	are used for training.
	Hong~\etal~\cite{hong2015decoupled} jointly train a network with image-level supervised images and  a few fully-annotated frames in the encoder-decoder framework. 
	The approaches in \cite{dai2015boxsup} and \cite{papandreou2015weakly} are generalized from the weakly-supervised to the semi-supervised setting for utilizing additional annotated image data.
	
	Different from the aforementioned methods, the proposed algorithm can leverage unlabeled images in model training, hence greatly alleviating the task of manual annotation.
	We treat the output of a fully convolutional discriminator as the supervisory signals, which compensate for the absence of image annotations and enable semi-supervised semantic segmentation.
	On the other hand, the proposed self-taught learning framework for segmentation is related to \cite{pathak2015constrained} where the prediction maps of unlabeled images are used as ground truth.
	However, in \cite{pathak2015constrained}, the prediction maps are refined by several hand-designed constraints before training, while we learn the selection criterion for self-taught learning based on the proposed adversarial network model.

	{\flushleft \bf Generative adversarial networks.}
	%
	Since the GAN framework with its theoretical foundation is proposed~\cite{gan}, it draws significant attention with several improvements in implementation~\cite{dcgan,denton2015deep, arjovsky2017wasserstein, mao2016multi, berthelot2017began} and applciations
	%
	including image generation~\cite{dcgan}, super-resolution~\cite{srgan,lai2017deep}, optical flow~\cite{lai2017semi}, object detection~\cite{fastarcnn}, domain adaptation~\cite{fcnsinthewild,Tsai_adaptseg_2018,Hoffman_cycada2017} and semantic segmentation~\cite{luc2016semantic, souly2017semi}.
	The work closest in scope to ours is the one proposed by \cite{luc2016semantic}, where the adversarial network is used to help the training process for semantic segmentation. 
	However, this method does not achieve substantial improvement over the baseline scheme 
	and does not tackle the semi-supervised setting.
	On the other hand, Souly~\etal \cite{souly2017semi} propose to generate adversarial examples using GAN for semi-supervised semantic segmentation. 
	%
	However, these generated examples may not be sufficiently close to real images to help the segmentation network since view synthesis from dense labels is still challenging.
	%
	%
	
	\vspace{-2mm}
	\section{Algorithm Overview}
	\vspace{-2mm}
	
\begin{figure}[t]
		\centering
		\includegraphics[width=0.95\linewidth]{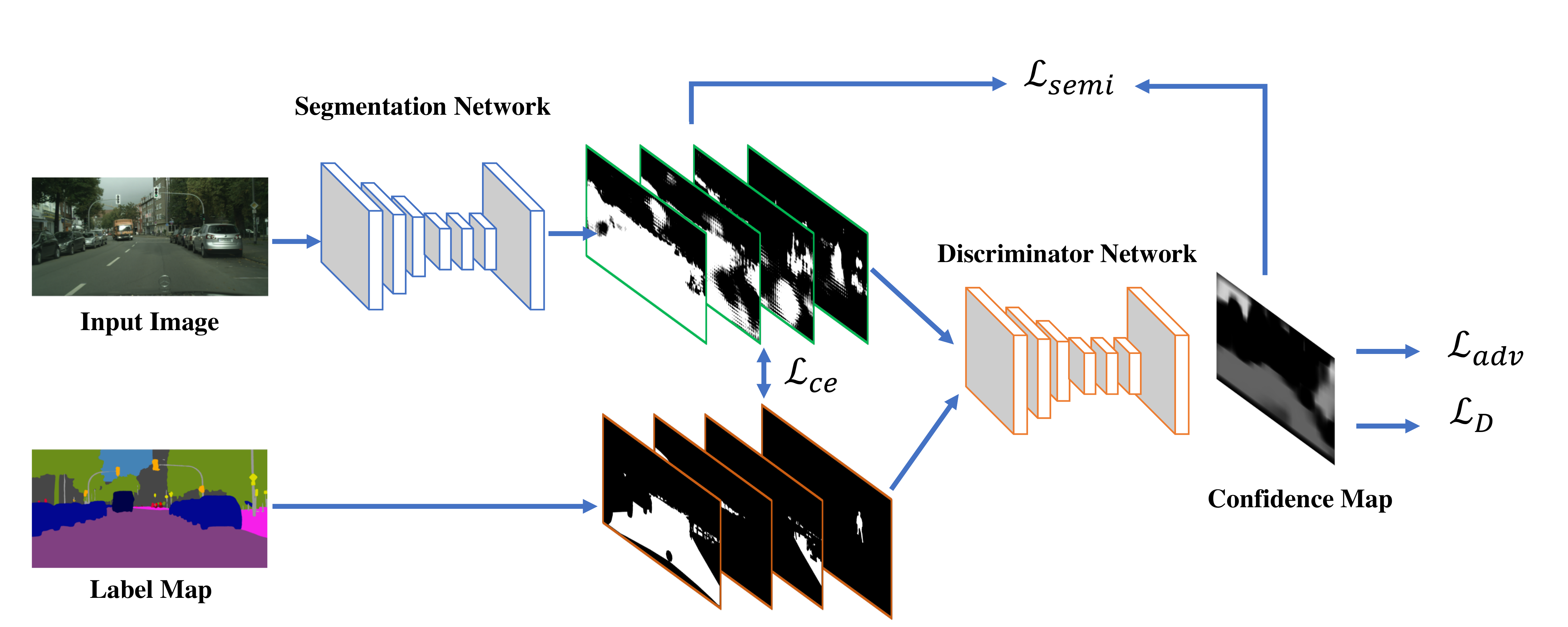}\\
		\vspace{-3mm}
		\caption{Overview of the proposed system for semi-supervised semantic segmentation. With a fully-convolution discriminator network trained using the loss $\mathcal{L}_{D}$, we optimize the segmentation network using three loss functions during the training process: cross-entropy loss $\mathcal{L}_{ce}$ on the segmentation ground truth, adversarial loss $\mathcal{L}_{adv}$ to fool the discriminator, and semi-supervised loss $\mathcal{L}_{semi}$ based on the confidence map, i.e., output of the discriminator.}
		\vspace{-0.5cm}
		\label{figure: semi_overview}
	\end{figure}

	Figure \ref{figure: semi_overview} shows the overview of the proposed algorithm.
	The proposed model consists of two modules: segmentation and discriminator networks.
	The former can be any network designed for semantic segmentation, e.g., FCN~\cite{fcn}, DeepLab~\cite{deeplab}, and DilatedNet~\cite{dilated}.
	Given an input image of dimension $H \times W \times 3$, the segmentation network outputs the class probability maps of size $H \times W \times C$, where $C$ is the number of semantic categories.

	Our discriminator network baed on a FCN, which takes class probability maps as the input, either from the segmentation network or ground truth label maps and then outputs spatial probability maps of size $H \times W \times 1$.
	Each pixel $p$ of the discriminator outputs map represents whether that pixel is sampled from the ground truth label ($p=1$) or the segmentation network ($p=0$).
	In contrast to the typical GAN discriminators which take fix-sized input images ($64 \times 64$ in most cases) and output a single probability value, we transform our discriminator to a fully-convolutional network that can take inputs of arbitrary sizes.
	More importantly, we show this transformation is essential for the proposed adversarial learning scheme.

	During the training process, we use both labeled and unlabeled images under the semi-supervised setting.
	When using the labeled data, the segmentation network is supervised by both the standard cross-entropy loss $\mathcal{L}_{ce}$ with the ground truth label map and the adversarial loss $\mathcal{L}_{adv}$ with the discriminator network.
	Note that we train the discriminator network only with the labeled data.
	%
	For the unlabeled data, we train the segmentation network with the proposed semi-supervised method. After obtaining the initial segmentation prediction of the unlabeled image from the segmentation network, we compute a confidence map by passing the segmentation prediction through the discriminator network.
	We in turn treat this confidence map as the supervisory signal using a self-taught 
	scheme to train the segmentation network with 
	a masked cross-entropy loss $\mathcal{L}_{semi}$.
	This confidence map indicates the quality of 
	the predicted segmented regions such that the segmentation network can trust during training.
	\vspace{-3mm}
	
	\section{Semi-Supervised Training with Adversarial Network}
	\vspace{-2mm}
	
	In this section, we present the proposed network architecture and learning schemes for the segmentation as well as discriminator modules.
	
	\subsection{Network Architecture}
	\vspace{-2mm}		
	{\flushleft \bf Segmentation network.}
	We adopt the DeepLab-v2~\cite{deeplab} framework with ResNet-101~\cite{resnet} model pre-trained on the ImageNet dataset~\cite{imagenet} and MSCOCO~\cite{lin2014microsoft} as our segmentation baseline network 
	(see Figure~\ref{figure: semi_overview}).
	However, we do not use the multi-scale fusion proposed in~\cite{deeplab} since it 
	would occupy all memory of a single GPU and make it impractical 
	to train the discriminator.
	Similar to the recent semantic segmentation method~\cite{deeplab,dilated}, we remove the last classification layer and modify the stride of the last two convolution layers from 2 to 1,
	thereby making the resolution of the output feature maps effectively $1/8$  
	of the input image size.
	To enlarge the receptive fields, we apply the dilated convolution~\cite{dilated} in conv4 and conv5 layers with strides of 2 and 4, respectively.
	%
	In addition, we use the Atrous Spatial Pyramid Pooling (ASPP) method~\cite{deeplab} in the last layer. 
	Finally, we apply an up-sampling layer along with the softmax output to match the size of the input image.
	
	\vspace{-2mm}	
	{\flushleft \bf Discriminator network.}
	We use the structure similar to \cite{dcgan} for the discriminator network.
	It consists of 5 convolution layers with $4 \times 4$ kernel and \{64, 128, 256, 512, 1\} channels in the stride of 2. 
	Each convolution layer is followed by a Leaky-ReLU~\cite{maas2013rectifier} parameterized by $0.2$ except the last layer.
	To transform the model into a fully convolutional network, an up-sampling layer is added to the last layer to rescale the output to the size of the input map.
	%
	%
	Note that we do not employ any batch-normalization layer~\cite{ioffe2015batch} as it only performs well when the batch size is sufficiently large. 
	
	\vspace{-2mm}
	\subsection{Loss Function}
	\label{section: training}
	
	Given an input image $\mathbf{X}_n$ of size $H \times W \times 3$, we denote the segmentation network by $S(\cdot)$ and the predicted probability map by $S(\mathbf{X}_n)$ of size $H \times W \times C$ where $C$ is the category number.
	We denote the fully convolutional discriminator by $D(\cdot)$ which takes a probability map 
	of  size $H \times W \times C$ and outputs a confidence map of size $H \times W \times 1$. 
	In the proposed method, there are two possible inputs to the discriminator network: 
	segmentation prediction $S(\mathbf{X}_n)$ or one-hot encoded ground truth vector 
	$\mathbf{Y}_n$.
	
	\vspace{-2mm}	
	{\flushleft \bf Discriminator network.}
	%
	To train the discriminator network, we minimize the spatial cross-entropy loss $\mathcal{L}_{D}$ with respect to two classes using:

	\begin{equation}
		\label{eqn: L_D}
		\mathcal{L}_{D} = - \sum_{h,w}{} (1-y_n) \log(1-D(S(\mathbf{X}_n))^{(h,w)}) + y_n \log(D(\mathbf{Y}_n)^{(h,w)}),
	\end{equation}
	where $y_n = 0$ if the sample is drawn from the segmentation network, and $y_n = 1$ if the sample is from the ground truth label.
	%
	In addition, $D(S(\mathbf{X}_n))^{(h,w)}$ is the confidence map of $\mathbf{X}$ 
	at location $(h, w)$, and $D(\mathbf{Y}_n)^{(h, w)}$ is defined similarly.
	%
	%
	To convert the ground truth label map with discrete labels to a $C$-channel probability map, we use the one-hot encoding scheme on the ground truth label map where $\mathbf{Y}_n^{(h,w,c)}$ takes value $1$ if pixel $\mathbf{X}_n^{(h,w)}$ belongs to class $c$, and $0$ otherwise.
	%
	
	One potential issue with the discriminator network is that it may easily distinguish whether the probability maps come from the ground truth by detecting the one-hot probability~\cite{luc2016semantic}. 
	However, we do not encounter this problem during the training phase.
	One reason is that we use a fully-convolutional scheme to predict spatial confidence, which increases the difficulty of learning the discriminator.
	In addition, we evaluate the \textit{Scale} scheme~\cite{luc2016semantic} 
	where the ground truth probability channel is slightly diffused to other channels according to the distribution of segmentation network output. 
	However, the results show no difference, and thus we do not adopt this scheme in this work.
	
	\vspace{-2mm}		
	{\flushleft \bf Segmentation network.}
	We train the segmentation network by minimizing a multi-task loss function:
	\begin{equation}
		\label{eqn: L_seg}
		\mathcal{L}_{seg} = \mathcal{L}_{ce} + \lambda_{adv} \mathcal{L}_{adv} +  \lambda_{semi} \mathcal{L}_{semi},
	\end{equation}
	where $\mathcal{L}_{ce}$, $\mathcal{L}_{adv}$, and $\mathcal{L}_{semi}$ denote the spatial multi-class cross entropy loss, adversarial loss, and semi-supervised loss, respectively.
	In~\eqref{eqn: L_seg}, $\lambda_{adv}$ and $\lambda_{semi}$ are two weights 
	for minimizing the proposed multi-task loss function.
	
	We first consider the scenario of using annotated data. Given an input image $\mathbf{X}_n$, its one-hot encoded ground truth $\mathbf{Y}_n$ and prediction result $S(\mathbf{X}_n)$, the cross-entropy loss is obtained by:
	\begin{equation}
		\mathcal{L}_{ce} = - \sum_{h,w}{} \sum_{c \in C}{} \mathbf{Y}_n^{(h,w,c)} \log(S(\mathbf{X}_n)^{(h,w,c)}).
	\end{equation}
	We use the adversarial learning process through the loss $\mathcal{L}_{adv}$ given a fully convolutional discriminator network $D(\cdot)$:
	\begin{equation}
		\mathcal{L}_{adv} = - \sum_{h,w}{} \log(D(S(\mathbf{X}_n))^{(h,w)}).
	\end{equation}
	%
	%
	With this loss, we train the segmentation network to fool the discriminator by maximizing the probability of the predicted results being generated from the ground truth distribution.
	
	\vspace{-2mm}		
	{\flushleft \bf Training with unlabeled data.}
	In this work, we consider the adversarial training under the semi-supervised setting.
	For unlabeled data, we do not apply $\mathcal{L}_{ce}$ since there is no ground truth annotation.
	The adversarial loss $\mathcal{L}_{adv}$ is still applicable as it only requires the discriminator network.
	However, we find that it is crucial to choose a smaller $\lambda_{adv}$ than the one 
	used for labeled data.
	It is because the adversarial loss may over-correct the prediction to fit the ground truth distribution without the cross entropy loss.
	
	In addition, we use the trained discriminator with unlabeled data within a  
	self-taught learning framework.
	%
	The main idea is that the trained discriminator can generate a confidence 
	map $D(S(\mathbf{X}_n))$ which can be used to infer the regions 
	sufficiently close to those from the ground truth distribution.
	We then binarize this confidence map with a threshold to highlight the trustworthy region. %
	Furthermore, the self-taught, one-hot encoded ground truth $\mathbf{\hat{Y}}_n$ is element-wise set with $\mathbf{\hat{Y}}_n^{(h,w,c^*)} = 1$ if $c^* = \arg \max_cS(\mathbf{X}_n)^{(h,w,c)}.$
	The resulting semi-supervised loss is defined by:
	\begin{equation}
		\label{eqn: L_semi}
		\mathcal{L}_{semi} = - \sum_{h,w}{} \sum_{c \in C}{} I(D(S(\mathbf{X}_n))^{(h,w)} > T_{semi}) \cdot \mathbf{\hat{Y}}_n^{(h,w,c)} \log(S(\mathbf{X}_n)^{(h,w,c)}),
	\end{equation}
	where $I(\cdot)$ is the indicator function and $T_{semi}$ is the threshold to control the sensitivity of the self-taught process.
	Note that during training we treat both the self-taught target $\mathbf{\hat{Y}}_n$ and the value of indicator function as constant, and thus (\ref{eqn: L_semi}) can be simply viewed as a masked spatial cross entropy loss.
	In practice, we find that this strategy works robustly with $T_{semi}$ ranging between $0.1$ and $0.3$.
	
	
	\vspace{-3mm}
	\section{Experimental Results}
	\vspace{-2mm}	
	{\flushleft \bf Implementation details.}
	We implement the proposed algorithm using the PyTorch framework.
	We train the proposed model on a single TitanX GPU with 12 GB memory.
	To train the segmentation network, we use the Stochastic Gradient Descent (SGD) optimization method, where the momentum is 0.9, and the weight decay is $10^{-4}$.
	The initial learning rate is set as $2.5 \times 10^{-4}$ and is decreased with polynomial decay with power of 0.9 as mentioned in \cite{deeplab}.
	For training the discriminator, we adopt Adam optimizer~\cite{kingma2014adam} with the learning rate $10^{-4}$ and the same polynomial decay as the segmentation network.
	For the hyper-parameters in the proposed method, $\lambda_{adv}$ is set as 0.01 and 0.001 when training with labeled and unlabeled data, respectively.
	We set $\lambda_{semi}$ as 0.1 and $T_{semi}$ as 0.2.
	
	For semi-supervised training, we randomly interleave labeled and unlabeled data while applying the training scheme described in Section \ref{section: training}.
	Note that, to prevent the model suffering from initial noisy masks and predictions, we start the semi-supervised learning after training for 5000 iterations with labeled data.
	We update both the segmentation network and discriminator network jointly. In each iteration, only the batch containing the ground truth data are used for training the discriminator.
	When randomly sampling partial labeled and unlabeled data from the datasets, we average several experiment results with different random seeds to ensure the evaluation robustness.
	The code and model are available at \url{https://github.com/hfslyc/AdvSemiSeg}.
	
	\vspace{-2mm}		
	{\flushleft \bf Evaluation datasets and metric.}
	In this work, we conduct experiments on two semantic segmentation datasets: PASCAL VOC 2012~\cite{pascal} and Cityscapes~\cite{cityscapes}. 
	%
	%
	On both datasets, we use the mean intersection-over-union (mean IU) as the evaluation metric.
	
	The PASCAL VOC 2012 dataset comprises 20 common objects with annotations on daily captured photos.
	%
	%
	%
	In addition, we utilize the extra annotated images from the Segmentation Boundaries Dataset (SBD) \cite{sbd} and obtain a set of total 10,582 training images. 
	%
	%
	We evaluate our models on the standard validation set with 1,449 images.
	During training, we use the random scaling and cropping operations with size $321 \times 321$.
	We train each model on the PASCAL VOC dataset for 20k iterations with batch size 10.
	
	The Cityscapes dataset contains 50 videos in driving scenes from which 
	2975, 500, 1525 images are extracted and annotated with 19 classes for training, validation, and testing, respectively.
	Each annotated frame is the $20$-{th} frame in a 30-frames snippet, where only these images with annotations are considered in the training process.
	We resize the input image to $512 \times 1024$ without any random cropping/scaling.
	We train each model on the Cityscapes dataset for 40k iterations with batch size 2.
	
	\begin{table}
		\begin{minipage}[t]{.49\linewidth}
			\scriptsize
			\caption{Results on the VOC 2012 \emph{val} set.}
			\vspace{1mm}
			\label{table: pascal}
			\centering
			\begin{tabular}{lcccc}
				\toprule
				& \multicolumn{4}{c}{Data Amount} \\
				Methods & 1/8 & 1/4 & 1/2 & Full \\
				\midrule
				FCN-8s~\cite{fcn} &N/A &N/A &N/A & 67.2 \\
				Dilation10~\cite{dilated} &N/A &N/A &N/A & 73.9 \\
				DeepLab-v2~\cite{deeplab} &N/A &N/A &N/A & 77.7 \\
				\midrule
				our baseline     & 66.0 & 68.3 & 69.8 & 73.6 \\
				baseline + $\mathcal{L}_{adv}$ & 67.6 & 71.0 & 72.6 & 74.9\\
				baseline + $\mathcal{L}_{adv}$ + $\mathcal{L}_{semi}$ & 69.5 & 72.1 & 73.8 & N/A \\
				
				\bottomrule
			\end{tabular}
		\end{minipage}
		\hfill
		\begin{minipage}[t]{.49\linewidth}
			\scriptsize
			\caption{Results on the Cityscapes \emph{val} set.}
			
			\label{table: cityscapes}
			\centering
			\begin{tabular}{lcccc}
				\toprule
				
				& \multicolumn{4}{c}{Data Amount} \\
				Methods & 1/8 & 1/4 & 1/2 & Full \\
				\midrule
				FCN-8s~\citep{fcn} &N/A &N/A &N/A & 65.3 \\
				Dilation10~\citep{dilated} &N/A &N/A &N/A & 67.1 \\
				DeepLab-v2~\citep{deeplab} &N/A &N/A &N/A & 70.4 \\
				\midrule
				our baseline     & 55.5 & 59.9 & 64.1 & 66.4 \\
				
				baseline + $\mathcal{L}_{adv}$ & 57.1 & 61.8 & 64.6 & 67.7 \\
				baseline + $\mathcal{L}_{adv}$ + $\mathcal{L}_{semi}$ & 58.8 & 62.3 & 65.7 & N/A \\
				
				\bottomrule
			\end{tabular}
		\end{minipage}
		\vspace{-5mm}
	\end{table}
	
	\vspace{-2mm}		
	{\flushleft \bf PASCAL VOC 2012.}
	Table \ref{table: pascal} shows the evaluation results on the PASCAL VOC 2012 dataset.
	To validate the semi-supervising scheme, we randomly sample 1/8, 1/4, 1/2 images as labeled data, and use the rest of training images as unlabeled data.
	We compare the proposed algorithm against the FCN~\cite{fcn}, Dilation10~\cite{dilated}, and DeepLab-v2~\cite{deeplab} methods.
	to demonstrate that our baseline model performs comparably with the state-of-the-art schemes.
	Note that our baseline model is equivalent to the DeepLab-v2 model without multi-scale fusion.
	The adversarial loss brings consistent performance improvement (from $1.6\%to 2.8\%$) over different amounts of training data.
	Incorporating the proposed semi-supervised learning scheme brings overall $3.5\%$ to $4.0\%$ improvement.
	%
	%
	%
	Figure \ref{fig: pascal} shows visual comparisons of the segmentation results generated by the proposed method. 
	We observe that the segmentation boundary achieves significant improvement when compared to the baseline model.
	
	\vspace{-2mm}		
	{\flushleft \bf Cityscapes.}
	Table \ref{table: cityscapes} shows evaluation results on the Cityscapes dataset.
	By applying the adversarial loss $\mathcal{L}_{adv}$, the model achieves $0.5\%$ to $1.9\%$ gain over the baseline model under the semi-supervised setting.
	This shows that our adversarial training scheme can encourage the segmentation network to learn the structural information from the ground truth distribution.
	Combining with  the adversarial learning and the proposed semi-supervised scheme, 
	the proposed algorithm achieves the performance gain of $1.6\%$ to $3.3\%$. 

	\vspace{-2mm}		
	{\flushleft \bf Comparisons with state-of-the-art methods.}
	Table \ref{table: adversarial_comp} shows comparisons with \cite{luc2016semantic} which utilizes adversarial learning for segmentation. 
	There are major differences between \cite{luc2016semantic} and our method
	in the adversarial learning processes. 
	First, we design a universal discriminator for various segmentation tasks, while \cite{luc2016semantic} utilizes one network structures for each dataset.
	Second, our discriminator does not require RGB images as additional inputs 
	but directly operates on the prediction maps from the segmentation network.
	%
	%
	%
	Table \ref{table: adversarial_comp} shows that our method achieves $1.2\%$ gain in mean IU over the method in \cite{luc2016semantic}.

	We present the results under the semi-supervised setting in Table \ref{table: semi_comp}.
	To compare with \cite{papandreou2015weakly} and \cite{souly2017semi}, our model is trained on the original PASCAL VOC 2012 train set (1,464 images) and use the SBD~\cite{sbd} set as unlabeled data.
	It is worth noticing that in \cite{papandreou2015weakly}, image-level labels are available for the SBD~\cite{sbd} set, and in \cite{souly2017semi} additional unlabeled images are generated through their generator during the training stage.
	
	\newcommand{\imgrow}[1]{
		\hspace{-4mm}
		\includegraphics[width = 0.18\linewidth, height=0.18\linewidth]{imgs/voc_results/img/#1.jpg}& \hspace{-4mm}
		\includegraphics[width = 0.18\linewidth, height=0.18\linewidth]{imgs/voc_results/voc_gt/#1.png} &\hspace{-4mm}
		\includegraphics[width = 0.18\linewidth, height=0.18\linewidth]{imgs/voc_results/voc_baseline/#1.png} &\hspace{-4mm}
		\includegraphics[width = 0.18\linewidth, height=0.18\linewidth]{imgs/voc_results/voc_Ladv/#1.png} &\hspace{-4mm}
		\includegraphics[width = 0.18\linewidth, height=0.18\linewidth]{imgs/voc_results/voc_Ladv+Lsemi/#1.png} \\
	}
	
	\begin{figure}[t]
		\scriptsize
		\centering
		\begin{tabular}{@{}ccccc@{}}
			\imgrow{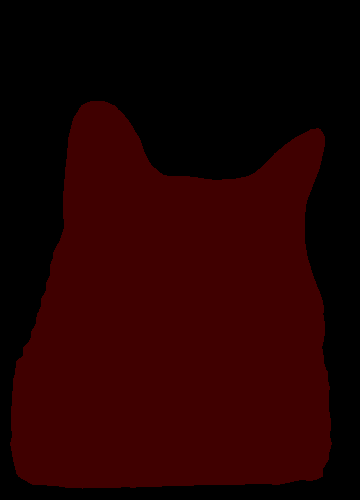}
			\imgrow{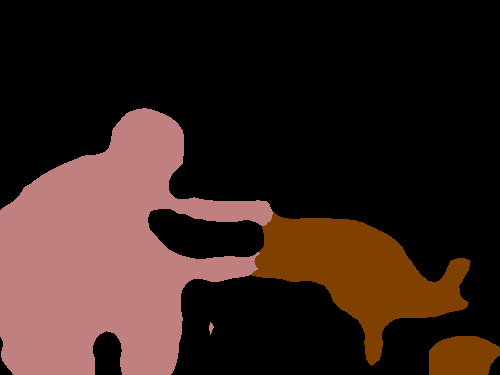}
			image & annotation & baseline & $+\mathcal{L}_{adv}$ & $+\mathcal{L}_{adv}+\mathcal{L}_{semi}$
		\end{tabular}
		\caption{Comparisons on the PASCAL VOC 2012 dataset using 1/2 labeled data.}
		\label{fig: pascal}
		\vspace{-2mm}
	\end{figure}
	
	\begin{table}
		\begin{minipage}[t]{.49\linewidth}
			\scriptsize
			\caption{Adversarial learning comparison with \cite{luc2016semantic} on the VOC 2012 \emph{validation} set.}
			\vspace{5mm}
			\label{table: adversarial_comp}
			\centering
			\begin{tabular}{lcc}
				\toprule
				& Baseline & Adversarial \\
				\midrule
				\cite{luc2016semantic} & 71.8 & 72.0\\
				ours & 73.6 & 74.9 \\
				
				\bottomrule
			\end{tabular}
		\end{minipage}
		\hfill
		\begin{minipage}[t]{.49\linewidth}
			
			\scriptsize
			\caption{Semi-supervised learning comparisons on the VOC 2012 \emph{validation} set without using additional labels of the SBD.}
			\vspace{1mm}
			\label{table: semi_comp}
			\centering
			\begin{tabular}{lccc}
				\toprule
				& Data & Fully- & Semi- \\
				& Amount& supervised & supervised \\
				\midrule
				\cite{papandreou2015weakly} & Full & 62.5 & 64.6 \\
				\cite{souly2017semi} & Full & 59.5 & 64.1\\
				ours & Full & 66.3 & 68.4 \\
				\midrule
				\cite{souly2017semi} & 30\% & 38.9 & 42.2\\
				ours & 30\% & 57.4 & 60.6 \\
				\bottomrule
			\end{tabular}
		\end{minipage}
		\vspace{-5mm}
	\end{table}

	\vspace{-2mm}	
	{\flushleft \bf Hyper-parameter analysis.}
	The proposed algorithm is governed by three hyper parameters: $\lambda_{adv}$ and $\lambda_{semi}$ for balancing the multi-task learning in (\ref{eqn: L_seg}), and $T_{semi}$ used to control the sensitivity in the semi-supervised learning described in (\ref{eqn: L_semi}).
	Table \ref{table: hyper} shows sensitivity analysis on hyper parameters using the PASCAL VOC dataset under the semi-supervised setting. 
	More analysis and results are provided in the supplementary material.
	%
	
\begin{figure}[t]
		\scriptsize
		\centering
		\begin{tabular}{@{}cccc@{}}

			\includegraphics[width = 0.18\linewidth, height=0.09\linewidth]{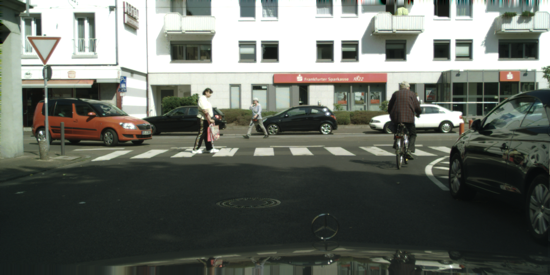}&
			\includegraphics[width = 0.18\linewidth, height=0.09\linewidth]{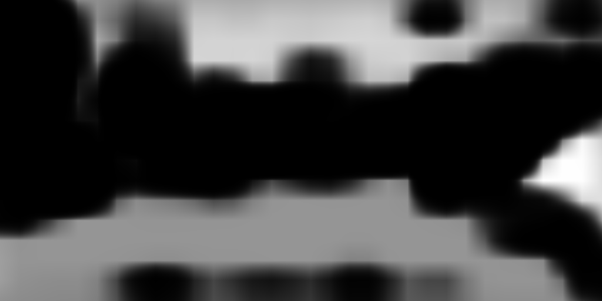}&
			\includegraphics[width = 0.18\linewidth, height=0.09\linewidth]{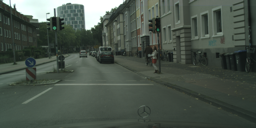}&
			\includegraphics[width = 0.18\linewidth, height=0.09\linewidth]{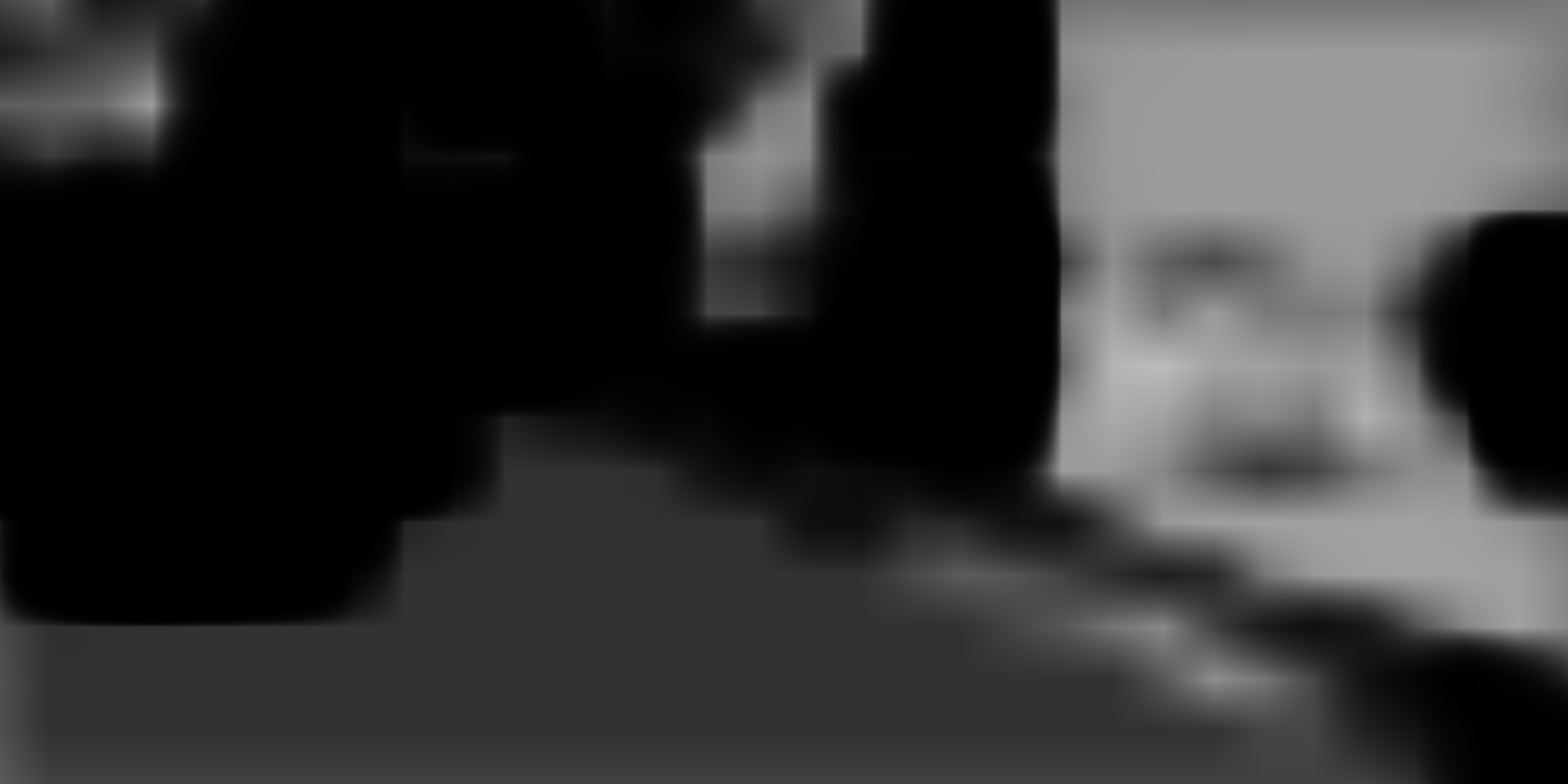} \\
			
			\includegraphics[width = 0.18\linewidth, height=0.09\linewidth]{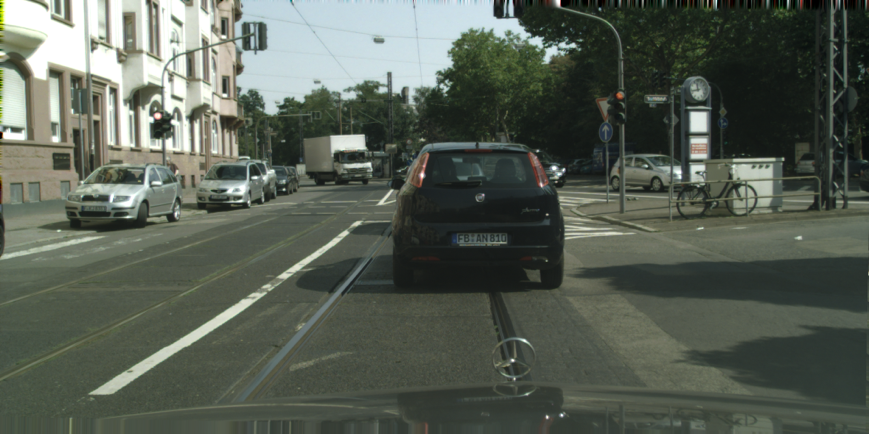}&
			\includegraphics[width = 0.18\linewidth, height=0.09\linewidth]{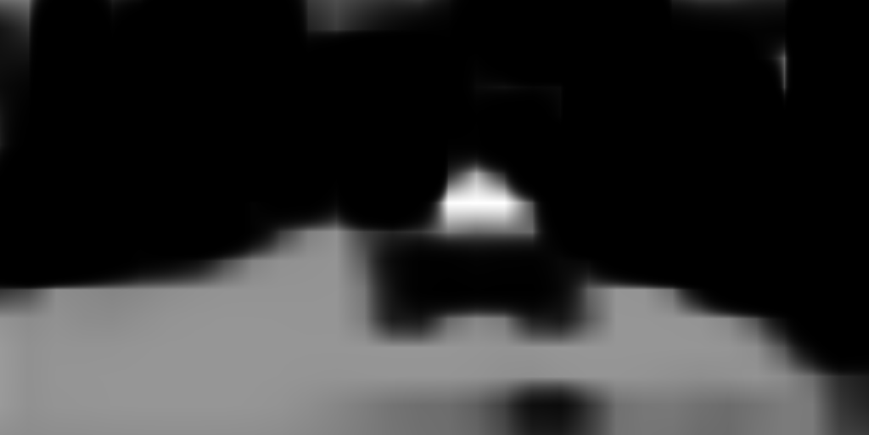}&
			\includegraphics[width = 0.18\linewidth, height=0.09\linewidth]{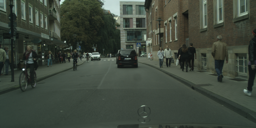}&
			\includegraphics[width = 0.18\linewidth, height=0.09\linewidth]{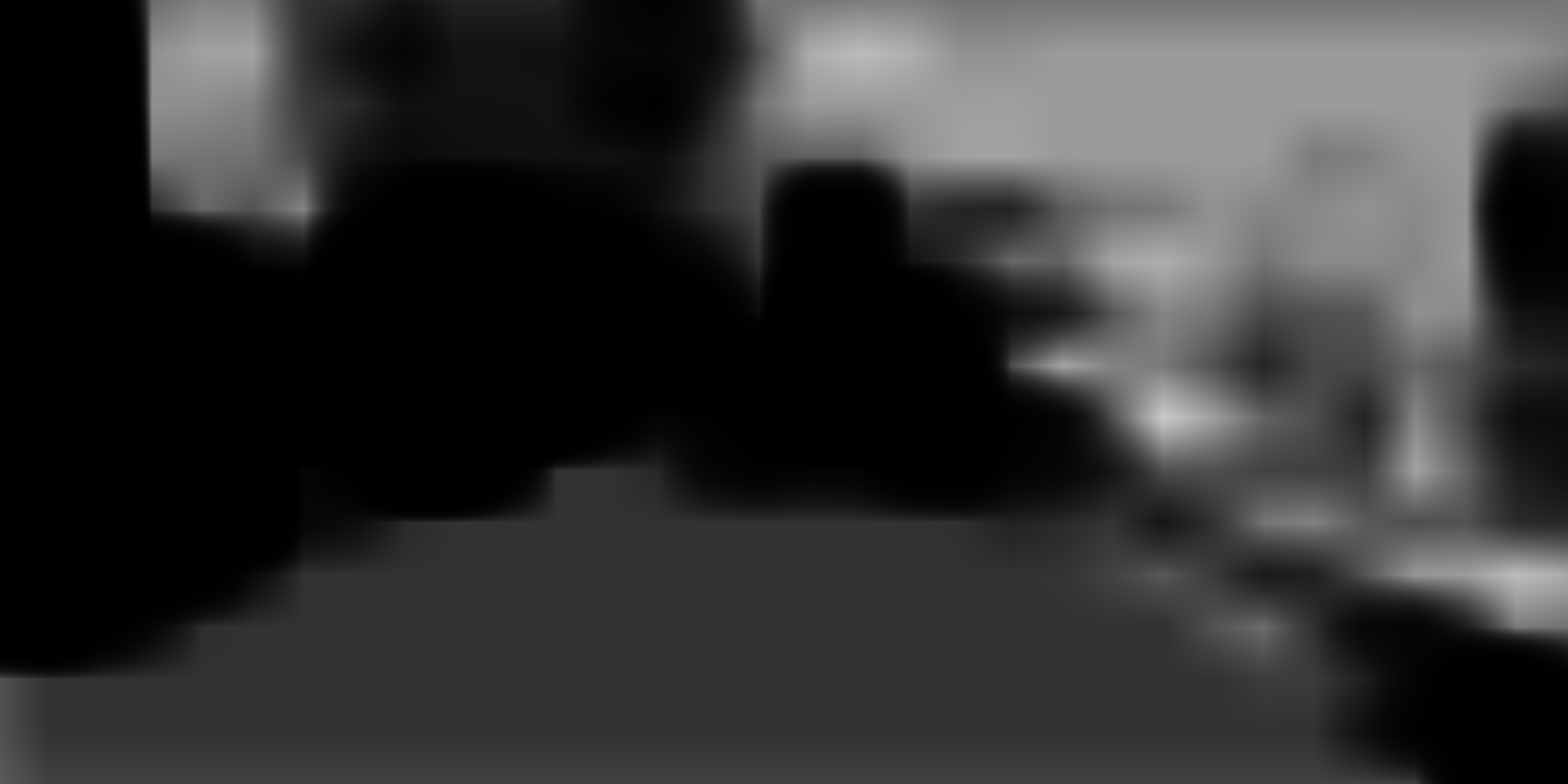} \\
			
			input image &  confidence map & input image &  confidence map \\
		\end{tabular}
		\caption{Visualization of the confidence maps. Given the prediction results generated by the segmentation network, the confidence maps are obtained from the discriminator. In the confidence maps, the brighter regions indicate that they are closer to the ground truth distribution, and we utilize these brighter regions for semi-supervised learning.}
		\label{fig: confidence_map}
		\vspace{-2mm}
	\end{figure}

	We first show comparisons of different values of $\lambda_{semi}$ with 1/8 amount of data under the semi-supervised setting. We set $\lambda_{adv} = 0.01$ and $T_{semi} = 0.2$ for the comparisons.
	%
	%
	Overall, the proposed method achieves the best mean IU of $69.5\%$ with $1.9\%$ gain.
	when $\lambda_{semi}$ is set to  $0.1$.
	Second, we perform the experiments with different values of $T_{semi}$ by setting $\lambda_{adv} = 0.01$ and $\lambda_{semi} = 0.1$ . 
	With higher $T_{semi}$, the proposed model only trusts regions of high structural similarity as the ground truth distribution.
	Overall, the proposed model achieves the best results when $T_{semi} = 0.2$
	and performs well for a wide range of $T_{semi}$ (0.1 to 0.3). 
	%
	When $T_{semi} =  0$, we trust all the pixel predictions in unlabeled images, which results in performance degradation.
	Figure \ref{fig: confidence_map} shows sample confidence maps from the predicted probability maps.
	
	\vspace{-3mm}		
	{\flushleft \bf Ablation study.}
	We present the ablation study of our proposed system in Table \ref{table: ablation} on the PASCAL VOC dataset.
	First, we examine the effect of using fully convolutional discriminator (FCD).
	To construct a discriminator that is not fully-convolutional, we replace the last convolution layer of the discriminator with a fully-connected layer that outputs a single neuron as in typical GAN models.
	%
	Without using FCD, the performance drops $1.0\%$ and $0.9\%$ with all and one-eighth data, respectively.
	This shows that the use of FCD is essential to adversarial learning.
	Second, we apply the semi-supervised learning method without the adversarial loss.
	The results show that the adversarial training on the labeled data is important to our semi-supervised scheme.
	If the segmentation network does not seek to fool the discriminator, the confidence maps generated by the discriminator would be meaningless, providing weaker supervisory signals.
	\vspace{-2mm}	
	
	\begin{table}[t]
		\begin{minipage}[c]{.49\linewidth}
			\caption{Hyper parameter analysis.}
			\scriptsize
			\vspace{2mm}
			\label{table: hyper}
			\centering
			\begin{tabular}{ccccc}
				\toprule
				Data Amount & $\lambda_{adv}$ & $\lambda_{semi}$ & $T_{semi}$ & Mean IU\\
				\midrule
				1/8&0.01&    0&    N/A    &67.6\\
				1/8    &0.01&    0.05&    0.2&    68.4\\
				1/8    &0.01&    0.1&    0.2&    69.5\\
				1/8    &0.01&    0.2&    0.2&    69.1\\
				\midrule
				1/8    &0.01&    0.1    &0&     67.2\\
				1/8    &0.01&    0.1    &0.1&    68.8\\
				1/8    &0.01&    0.1&    0.2&    69.5\\
				1/8    &0.01&    0.1    &0.3&    69.2\\
				1/8    &0.01&    0.1    &1.0&    67.6\\
				\bottomrule
			\end{tabular}
		\end{minipage}
		\hfill
		\begin{minipage}[c]{.49\linewidth}
			\scriptsize
			\caption{Ablation study of the proposed method on the PASCAL VOC dataset.}
			\vspace{2mm}
			\label{table: ablation}
			\centering
			\begin{tabular}{ccc|cc}
				\toprule
				& & & \multicolumn{2}{c}{Data Amount} \\
				
				$\mathcal{L}_{adv}$ & $\mathcal{L}_{semi}$ & FCD         & 1/8  & Full \\
				\midrule
				&              &             & 66.0  & 73.6 \\
				\checkmark  &            &  \checkmark & 67.6  & 74.9 \\
				\checkmark  &            &             & 66.6  & 74.0  \\
				\midrule
				& \checkmark &  \checkmark & 65.7  & N/A \\
				\checkmark  & \checkmark &  \checkmark & 69.5 & N/A \\
				\bottomrule
			\end{tabular}
		\end{minipage}
		\vspace{-5mm}
	\end{table}
	
	\vspace{-4mm}
	\section{Conclusions}
	\vspace{-3mm}
	In this work, we propose an adversarial learning scheme for semi-supervised semantic segmentation.
	We train a discriminator network to enhance the segmentation network with both labeled and unlabeled data. With labeled data, the adversarial loss for the segmentation network is designed to learn higher order structural information without post-processing.
	For unlabeled data, the confidence maps generated by the discriminator network act as the self-taught signal for refining the segmentation network.
	Extensive experiments on the PASCAL VOC 2012 and Cityscapes datasets validate the effectiveness of the proposed algorithm.
	
	{\flushleft {\bf Acknowledgments.}}
	W.-C. Hung is supported in part by the NSF CAREER Grant \#1149783, gifts from Adobe and NVIDIA.
	
	\bibliography{semi_seg}

\begin{thebibliography}{45}
\providecommand{\natexlab}[1]{#1}
\providecommand{\url}[1]{\texttt{#1}}
\expandafter\ifx\csname urlstyle\endcsname\relax
  \providecommand{\doi}[1]{doi: #1}\else
  \providecommand{\doi}{doi: \begingroup \urlstyle{rm}\Url}\fi

\bibitem[Arjovsky et~al.(2017)Arjovsky, Chintala, and
  Bottou]{arjovsky2017wasserstein}
Martin Arjovsky, Soumith Chintala, and L{\'e}on Bottou.
\newblock Wasserstein gan.
\newblock \emph{arXiv preprint arXiv:1701.07875}, 2017.

\bibitem[Bearman et~al.(2016)Bearman, Russakovsky, Ferrari, and
  Fei-Fei]{bearman2016s}
Amy Bearman, Olga Russakovsky, Vittorio Ferrari, and Li~Fei-Fei.
\newblock What’s the point: Semantic segmentation with point supervision.
\newblock In \emph{ECCV}, 2016.

\bibitem[Berthelot et~al.(2017)Berthelot, Schumm, and Metz]{berthelot2017began}
David Berthelot, Tom Schumm, and Luke Metz.
\newblock Began: Boundary equilibrium generative adversarial networks.
\newblock \emph{arXiv preprint arXiv:1703.10717}, 2017.

\bibitem[Chen et~al.(2017)Chen, Papandreou, Kokkinos, Murphy, and
  Yuille]{deeplab}
Liang-Chieh Chen, George Papandreou, Iasonas Kokkinos, Kevin Murphy, and Alan~L
  Yuille.
\newblock Deeplab: Semantic image segmentation with deep convolutional nets,
  atrous convolution, and fully connected crfs.
\newblock In \emph{TPAMI}, 2017.

\bibitem[Cordts et~al.(2016)Cordts, Omran, Ramos, Rehfeld, Enzweiler, Benenson,
  Franke, Roth, and Schiele]{cityscapes}
Marius Cordts, Mohamed Omran, Sebastian Ramos, Timo Rehfeld, Markus Enzweiler,
  Rodrigo Benenson, Uwe Franke, Stefan Roth, and Bernt Schiele.
\newblock The cityscapes dataset for semantic urban scene understanding.
\newblock In \emph{CVPR}, 2016.

\bibitem[Dai et~al.(2015)Dai, He, and Sun]{dai2015boxsup}
Jifeng Dai, Kaiming He, and Jian Sun.
\newblock Boxsup: Exploiting bounding boxes to supervise convolutional networks
  for semantic segmentation.
\newblock In \emph{ICCV}, 2015.

\bibitem[Deng et~al.(2009)Deng, Dong, Socher, Li, Li, and Fei-Fei]{imagenet}
Jia Deng, Wei Dong, Richard Socher, Li-Jia Li, Kai Li, and Li~Fei-Fei.
\newblock Imagenet: A large-scale hierarchical image database.
\newblock In \emph{CVPR}, 2009.

\bibitem[Denton et~al.(2015)Denton, Chintala, Fergus, et~al.]{denton2015deep}
Emily~L Denton, Soumith Chintala, Rob Fergus, et~al.
\newblock Deep generative image models using a laplacian pyramid of adversarial
  networks.
\newblock In \emph{NIPS}, 2015.

\bibitem[Everingham et~al.(2010)Everingham, Van~Gool, Williams, Winn, and
  Zisserman]{pascal}
Mark Everingham, Luc Van~Gool, Christopher~KI Williams, John Winn, and Andrew
  Zisserman.
\newblock The pascal visual object classes (voc) challenge.
\newblock In \emph{IJCV}, 2010.

\bibitem[Goodfellow et~al.(2014)Goodfellow, Pouget-Abadie, Mirza, Xu,
  Warde-Farley, Ozair, Courville, and Bengio]{gan}
Ian Goodfellow, Jean Pouget-Abadie, Mehdi Mirza, Bing Xu, David Warde-Farley,
  Sherjil Ozair, Aaron Courville, and Yoshua Bengio.
\newblock Generative adversarial nets.
\newblock In \emph{NIPS}, 2014.

\bibitem[Hariharan et~al.(2011)Hariharan, Arbel{\'a}ez, Bourdev, Maji, and
  Malik]{sbd}
Bharath Hariharan, Pablo Arbel{\'a}ez, Lubomir Bourdev, Subhransu Maji, and
  Jitendra Malik.
\newblock Semantic contours from inverse detectors.
\newblock In \emph{ICCV}, 2011.

\bibitem[He et~al.(2016)He, Zhang, Ren, and Sun]{resnet}
Kaiming He, Xiangyu Zhang, Shaoqing Ren, and Jian Sun.
\newblock Deep residual learning for image recognition.
\newblock In \emph{CVPR}, 2016.

\bibitem[Hoffman et~al.(2016)Hoffman, Wang, Yu, and Darrell]{fcnsinthewild}
Judy Hoffman, Dequan Wang, Fisher Yu, and Trevor Darrell.
\newblock Fcns in the wild: Pixel-level adversarial and constraint-based
  adaptation.
\newblock In \emph{arXiv preprint arXiv:1612.02649}, 2016.

\bibitem[Hoffman et~al.(2018)Hoffman, Tzeng, Park, Zhu, Isola, Saenko, Efros,
  and Darrell]{Hoffman_cycada2017}
Judy Hoffman, Eric Tzeng, Taesung Park, Jun-Yan Zhu, Phillip Isola, Kate
  Saenko, Alexei~A. Efros, and Trevor Darrell.
\newblock Cycada: Cycle consistent adversarial domain adaptation.
\newblock In \emph{ICML}, 2018.

\bibitem[Hong et~al.(2015)Hong, Noh, and Han]{hong2015decoupled}
Seunghoon Hong, Hyeonwoo Noh, and Bohyung Han.
\newblock Decoupled deep neural network for semi-supervised semantic
  segmentation.
\newblock In \emph{NIPS}, 2015.

\bibitem[Hong et~al.(2017)Hong, Yeo, Kwak, Lee, and Han]{hong2017weakly}
Seunghoon Hong, Donghun Yeo, Suha Kwak, Honglak Lee, and Bohyung Han.
\newblock Weakly supervised semantic segmentation using web-crawled videos.
\newblock In \emph{CVPR}, 2017.

\bibitem[Hung et~al.(2017)Hung, Tsai, Shen, Lin, Sunkavalli, Lu, and
  Yang]{hung2017scene}
Wei-Chih Hung, Yi-Hsuan Tsai, Xiaohui Shen, Zhe Lin, Kalyan Sunkavalli, Xin Lu,
  and Ming-Hsuan Yang.
\newblock Scene parsing with global context embedding.
\newblock In \emph{ICCV}, 2017.

\bibitem[Ioffe and Szegedy(2015)]{ioffe2015batch}
Sergey Ioffe and Christian Szegedy.
\newblock Batch normalization: Accelerating deep network training by reducing
  internal covariate shift.
\newblock In \emph{ICML}, 2015.

\bibitem[Khoreva et~al.(2017)Khoreva, Benenson, Hosang, Hein, and
  Schiele]{khoreva_CVPR17}
A.~Khoreva, R.~Benenson, J.~Hosang, M.~Hein, and B.~Schiele.
\newblock Simple does it: Weakly supervised instance and semantic segmentation.
\newblock In \emph{CVPR}, 2017.

\bibitem[Kingma and Ba(2014)]{kingma2014adam}
Diederik Kingma and Jimmy Ba.
\newblock Adam: A method for stochastic optimization.
\newblock In \emph{arXiv preprint arXiv:1412.6980}, 2014.

\bibitem[Krizhevsky et~al.(2012)Krizhevsky, Sutskever, and Hinton]{alexnet}
Alex Krizhevsky, Ilya Sutskever, and Geoffrey~E Hinton.
\newblock Imagenet classification with deep convolutional neural networks.
\newblock In \emph{NIPS}, 2012.

\bibitem[Lai et~al.(2017{\natexlab{a}})Lai, Huang, Ahuja, and
  Yang]{lai2017deep}
Wei-Sheng Lai, Jia-Bin Huang, Narendra Ahuja, and Ming-Hsuan Yang.
\newblock Deep laplacian pyramid networks for fast and accurate
  superresolution.
\newblock In \emph{CVPR}, 2017{\natexlab{a}}.

\bibitem[Lai et~al.(2017{\natexlab{b}})Lai, Huang, and Yang]{lai2017semi}
Wei-Sheng Lai, Jia-Bin Huang, and Ming-Hsuan Yang.
\newblock Semi-supervised learning for optical flow with generative adversarial
  networks.
\newblock In \emph{NIPS}, 2017{\natexlab{b}}.

\bibitem[Ledig et~al.(2016)Ledig, Theis, Husz{\'a}r, Caballero, Cunningham,
  Acosta, Aitken, Tejani, Totz, Wang, et~al.]{srgan}
Christian Ledig, Lucas Theis, Ferenc Husz{\'a}r, Jose Caballero, Andrew
  Cunningham, Alejandro Acosta, Andrew Aitken, Alykhan Tejani, Johannes Totz,
  Zehan Wang, et~al.
\newblock Photo-realistic single image super-resolution using a generative
  adversarial network.
\newblock In \emph{arXiv preprint arXiv:1609.04802}, 2016.

\bibitem[Lin et~al.(2016)Lin, Shen, van~dan Hengel, and Reid]{piecewise}
Guosheng Lin, Chunhua Shen, Anton van~dan Hengel, and Ian Reid.
\newblock {Efficient piecewise training of deep structured models for semantic
  segmentation}.
\newblock In \emph{CVPR}, 2016.

\bibitem[Lin et~al.(2014)Lin, Maire, Belongie, Hays, Perona, Ramanan,
  Doll{\'a}r, and Zitnick]{lin2014microsoft}
Tsung-Yi Lin, Michael Maire, Serge Belongie, James Hays, Pietro Perona, Deva
  Ramanan, Piotr Doll{\'a}r, and C~Lawrence Zitnick.
\newblock Microsoft coco: Common objects in context.
\newblock In \emph{ECCV}, 2014.

\bibitem[Liu et~al.(2015)Liu, Li, Luo, Loy, and Tang]{deepparsing}
Ziwei Liu, Xiaoxiao Li, Ping Luo, Chen~Change Loy, and Xiaoou Tang.
\newblock {Semantic Image Segmentation via Deep Parsing Network}.
\newblock In \emph{ICCV}, 2015.

\bibitem[Long et~al.(2015)Long, Shelhamer, and Darrell]{fcn}
Jonathan Long, Evan Shelhamer, and Trevor Darrell.
\newblock {Fully convolutional networks for semantic segmentation}.
\newblock In \emph{CVPR}, 2015.

\bibitem[Luc et~al.(2016)Luc, Couprie, Chintala, and Verbeek]{luc2016semantic}
Pauline Luc, Camille Couprie, Soumith Chintala, and Jakob Verbeek.
\newblock Semantic segmentation using adversarial networks.
\newblock In \emph{NIPS Workshop on Adversarial Training}, 2016.

\bibitem[Maas et~al.(2013)Maas, Hannun, and Ng]{maas2013rectifier}
Andrew~L Maas, Awni~Y Hannun, and Andrew~Y Ng.
\newblock Rectifier nonlinearities improve neural network acoustic models.
\newblock In \emph{ICML}, 2013.

\bibitem[Mao et~al.(2016)Mao, Li, Xie, Lau, and Wang]{mao2016multi}
Xudong Mao, Qing Li, Haoran Xie, Raymond~YK Lau, and Zhen Wang.
\newblock Multi-class generative adversarial networks with the l2 loss
  function.
\newblock \emph{arXiv preprint arXiv:1611.04076}, 2016.

\bibitem[Mottaghi et~al.(2014)Mottaghi, Chen, Liu, Cho, Lee, Fidler, Urtasun,
  and Yuille]{pascal_context}
Roozbeh Mottaghi, Xianjie Chen, Xiaobai Liu, Nam-Gyu Cho, Seong-Whan Lee, Sanja
  Fidler, Raquel Urtasun, and Alan Yuille.
\newblock {The Role of Context for Object Detection and Semantic Segmentation
  in the Wild}.
\newblock In \emph{CVPR}, 2014.

\bibitem[Papandreou et~al.(2015)Papandreou, Chen, Murphy, and
  Yuille]{papandreou2015weakly}
George Papandreou, Liang-Chieh Chen, Kevin Murphy, and Alan~L Yuille.
\newblock Weakly-and semi-supervised learning of a dcnn for semantic image
  segmentation.
\newblock In \emph{ICCV}, 2015.

\bibitem[Pathak et~al.(2015{\natexlab{a}})Pathak, Krahenbuhl, and
  Darrell]{pathak2015constrained}
Deepak Pathak, Philipp Krahenbuhl, and Trevor Darrell.
\newblock Constrained convolutional neural networks for weakly supervised
  segmentation.
\newblock In \emph{ICCV}, 2015{\natexlab{a}}.

\bibitem[Pathak et~al.(2015{\natexlab{b}})Pathak, Shelhamer, Long, and
  Darrell]{pathak2014fully}
Deepak Pathak, Evan Shelhamer, Jonathan Long, and Trevor Darrell.
\newblock Fully convolutional multi-class multiple instance learning.
\newblock In \emph{ICLR}, 2015{\natexlab{b}}.

\bibitem[Pinheiro and Collobert(2015)]{pinheiro2015weakly}
Pedro~O Pinheiro and Ronan Collobert.
\newblock Weakly supervised semantic segmentation with convolutional networks.
\newblock In \emph{CVPR}, 2015.

\bibitem[Qi et~al.(2016)Qi, Liu, Shi, Zhao, and Jia]{qi2016augmented}
Xiaojuan Qi, Zhengzhe Liu, Jianping Shi, Hengshuang Zhao, and Jiaya Jia.
\newblock Augmented feedback in semantic segmentation under image level
  supervision.
\newblock In \emph{ECCV}, 2016.

\bibitem[Radford et~al.(2016)Radford, Metz, and Chintala]{dcgan}
Alec Radford, Luke Metz, and Soumith Chintala.
\newblock Unsupervised representation learning with deep convolutional
  generative adversarial networks.
\newblock In \emph{ICLR}, 2016.

\bibitem[Simonyan and Zisserman(2015)]{vgg}
Karen Simonyan and Andrew Zisserman.
\newblock Very deep convolutional networks for large-scale image recognition.
\newblock In \emph{ICLR}, 2015.

\bibitem[Souly et~al.(2017)Souly, Spampinato, and Shah]{souly2017semi}
Nasim Souly, Concetto Spampinato, and Mubarak Shah.
\newblock Semi and weakly supervised semantic segmentation using generative
  adversarial network.
\newblock In \emph{ICCV}, 2017.

\bibitem[Tsai et~al.(2018)Tsai, Hung, Schulter, Sohn, Yang, and
  Chandraker]{Tsai_adaptseg_2018}
Yi-Hsuan Tsai, Wei-Chih Hung, Samuel Schulter, Kihyuk Sohn, Ming-Hsuan Yang,
  and Manmohan Chandraker.
\newblock Learning to adapt structured output space for semantic segmentation.
\newblock In \emph{CVPR}, 2018.

\bibitem[Wang et~al.(2017)Wang, Shrivastava, and Gupta]{fastarcnn}
Xiaolong Wang, Abhinav Shrivastava, and Abhinav Gupta.
\newblock A-fast-rcnn: Hard positive generation via adversary for object
  detection.
\newblock In \emph{CVPR}, 2017.

\bibitem[Yu and Koltun(2016)]{dilated}
Fisher Yu and Vladlen Koltun.
\newblock Multi-scale context aggregation by dilated convolutions.
\newblock In \emph{ICLR}, 2016.

\bibitem[Zheng et~al.(2015)Zheng, Jayasumana, Romera-Paredes, Vineet, Su, Du,
  Huang, and Torr]{crfasrnn}
Shuai Zheng, Sadeep Jayasumana, Bernardino Romera-Paredes, Vibhav Vineet,
  Zhizhong Su, Dalong Du, Chang Huang, and Philip~HS Torr.
\newblock Conditional random fields as recurrent neural networks.
\newblock In \emph{ICCV}, 2015.

\bibitem[Zhou et~al.(2017)Zhou, Zhao, Puig, Fidler, Barriuso, and
  Torralba]{zhou2016semantic}
Bolei Zhou, Hang Zhao, Xavier Puig, Sanja Fidler, Adela Barriuso, and Antonio
  Torralba.
\newblock Semantic understanding of scenes through the ade20k dataset.
\newblock In \emph{CVPR}, 2017.

\end{thebibliography}
	
	\clearpage
	\appendix
	
	\section{Pixel Accuracy in Semi-Supervised Learning}
	
	In Table~\ref{table:secmi_acc}, we show the average segmentation accuracy with respect to the number of selected pixels based on different threshold values of $T_{semi}$ as in (5) of the paper on the Cityscapes dataset. 
	With a higher $T_{semi}$, the discriminator outputs are more confident (similar to ground truth label distributions) and lead to more accurate pixel predictions.
	Also, as a trade-off, the higher threshold ($T_{semi}$), the fewer pixels are selected for back-propagation. This trade-off could also be observed in Table 5 of the paper.
	
	\begin{table}[h]
		\caption{Selected pixel accuracy.}
		\centering
		\label{table:secmi_acc}
		\begin{tabular}{c|c|c}
			\toprule
			$T_{semi}$ & Selected Pixels (\%) & Accuracy \\
			\midrule
			0 & 100\% & 92.65\%\\
			0.1 & 36\% & 99.84\% \\
			0.2 & 31\% & 99.91\% \\
			0.3 & 27\% & 99.94\% \\ 
			\bottomrule
		\end{tabular}
		
	\end{table}
	\section{Additional Hyper-parameter Analysis}
	In Table~\ref{table: hyper}, we show the complete hyper-parameter analysis.
	In addition to the analysis of $\lambda_{semi}$ and $T_{semi}$ in Table 5 of the paper, we show that the proposed adversarial learning is also robust to different values of $\lambda_{adv}$.

	\begin{table}[h]
		\caption{Hyper parameter analysis.}
		\small
		\vspace{2mm}
		\label{table: hyper}
		\centering
		\begin{tabular}{ccccc}
			\toprule
			Data Amount & $\lambda_{adv}$ & $\lambda_{semi}$ & $T_{semi}$ & Mean IU\\
			\midrule
			Full & 0 & 0 & N/A & 73.6 \\
			Full&0.005&	0&	N/A	&74.0\\
			Full&0.01&	0&	N/A	&74.9\\
			Full&0.02&	0&	N/A	&74.6\\
			Full&0.04&	0&	N/A	&74.1\\
			Full&0.05& 0 & N/A & 73.0 \\
			\midrule
			1/8&0.01&	0&	N/A	&67.6\\
			1/8	&0.01&	0.05&	0.2&	68.4\\
			1/8	&0.01&	0.1&	0.2&	69.5\\
			1/8	&0.01&	0.2&	0.2&	69.1\\
			\midrule
			1/8	&0.01&	0.1	&0&	 67.2\\
			1/8	&0.01&	0.1	&0.1&	68.8\\
			1/8	&0.01&	0.1&	0.2&	69.5\\
			1/8	&0.01&	0.1	&0.3&	69.2\\
			1/8	&0.01&	0.1	&1.0&	67.6\\
			\bottomrule
		\end{tabular}
	\end{table}

	\section{Training Parameters}
	
	In Table~\ref{table:train_param}, we show the training parameters for both datasets.
	We use the PyTorch implementation, and we will release our code and models for the public.
	\vspace{1cm}
	\begin{table}[h]
		\caption{Training parameters.}
		\centering
		\small
		\label{table:train_param}
		\begin{tabular}{l|c|c}
			\toprule
			Parameter & Cityscaps & PASCAL VOC \\
			\midrule
			Trained iterations & 40,000 & 20,000 \\
			Learning rate & 2.5e-4 &  2.5e-4\\
			Learning rate (D) & 1e-4 &  1e-4\\
			Polynomial decay & 0.9 & 0.9 \\
			Momentum & 0.9 & 0.9 \\
			Optimizer & SGD & SGD \\
			Optimizer (D) & Adam & Adam \\
			Nesterov & True & True \\ 
			Batch size & 2 & 10 \\
			Weight decay & 0.0001 & 0.0001 \\
			Crop size & 512x1024 & 321x321 \\
			Random scale & No & Yes \\

			\bottomrule
		\end{tabular}
	\end{table}
	
	\section{Additional Qualitative Results}
	
	In Figure \ref{fig: pascal1}-\ref{fig: pascal2}, we show additional qualitative comparisons with the models using half training data of the PSCAL VOC dataset.
	In Figure \ref{fig: CS1}, we show more qualitative comparisons with the models using half training data of the Cityscapes dataset.
	The results show that both the adversarial learning and the semi-supervised training scheme improve the segmentation quality.
	
	\newcommand{\imgrowvoc}[1]{
		\hspace{-4mm}
		\includegraphics[width = 0.18\linewidth, height=0.18\linewidth]{imgs/voc_results/img/#1.jpg}& \hspace{-4mm}
		\includegraphics[width = 0.18\linewidth, height=0.18\linewidth]{imgs/voc_results/voc_gt/#1.png} &\hspace{-4mm}
		\includegraphics[width = 0.18\linewidth, height=0.18\linewidth]{imgs/voc_results/voc_baseline/#1.png} &\hspace{-4mm}
		\includegraphics[width = 0.18\linewidth, height=0.18\linewidth]{imgs/voc_results/voc_Ladv/#1.png} &\hspace{-4mm}
		\includegraphics[width = 0.18\linewidth, height=0.18\linewidth]{imgs/voc_results/voc_Ladv+Lsemi/#1.png} \\
	}
	
	\begin{figure}[t]
		\scriptsize
		\centering
		\begin{tabular}{@{}ccccc@{}}
			\imgrowvoc{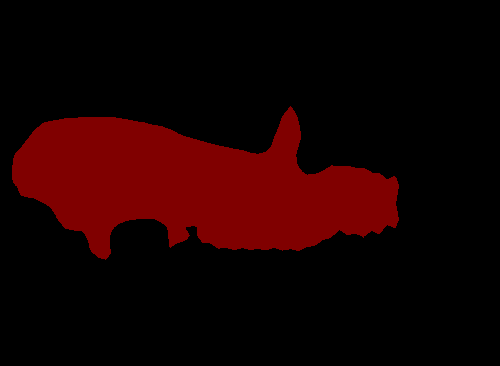}
			\imgrowvoc{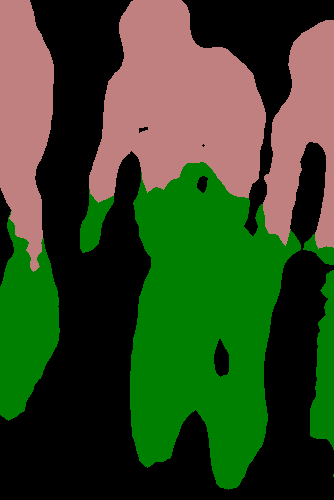}
			\imgrowvoc{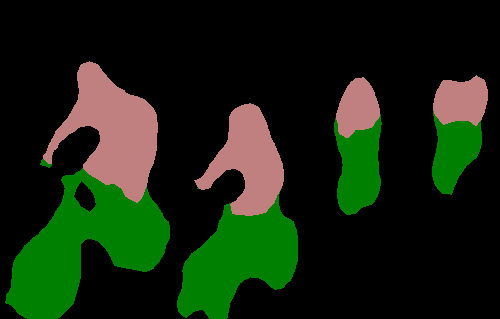}
			\imgrowvoc{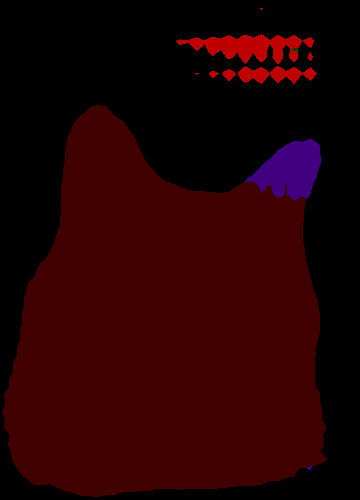}
			\imgrowvoc{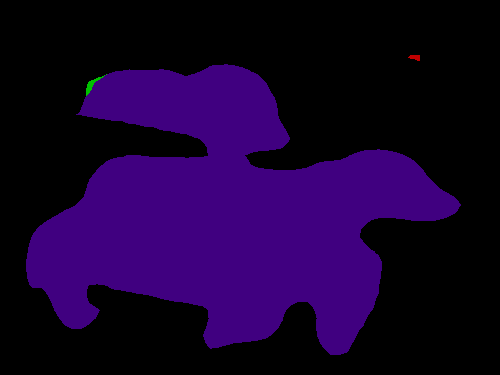}
			\imgrowvoc{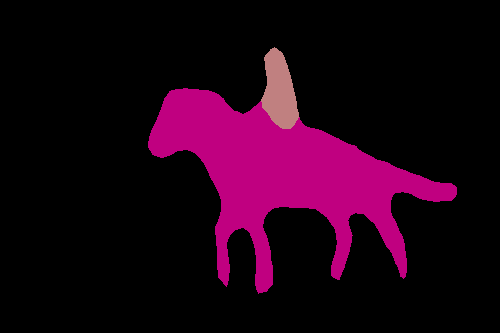}
			\imgrowvoc{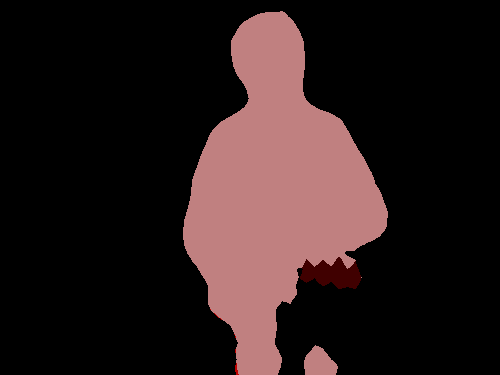}
			
			image & annotation & baseline & $+\mathcal{L}_{adv}$ & $+\mathcal{L}_{adv}+\mathcal{L}_{semi}$
		\end{tabular}
		\caption{Comparisons on the PASCAL VOC dataset using 1/2 training data.}
		\label{fig: pascal1}
	\end{figure}

	\begin{figure}[t]
		\scriptsize
		\centering
		\begin{tabular}{@{}ccccc@{}}
			\imgrowvoc{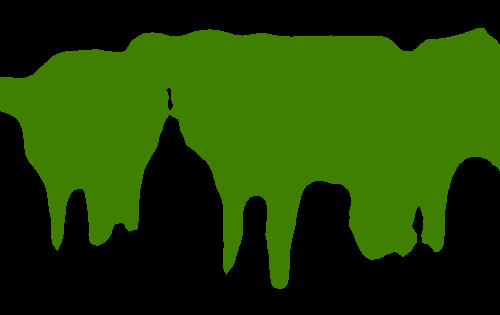}
			\imgrowvoc{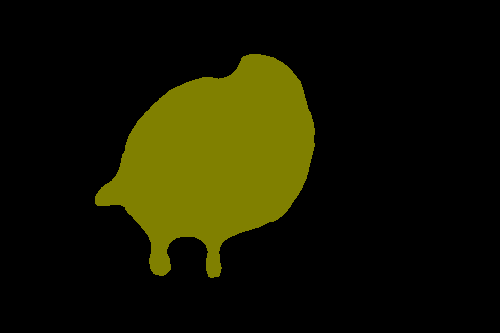}
			\imgrowvoc{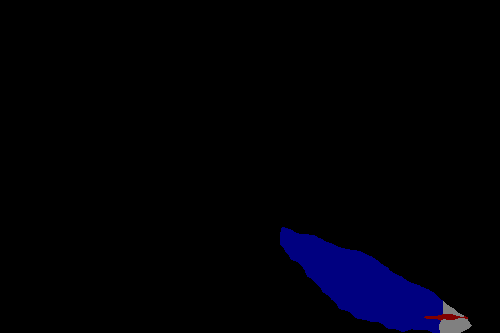}
			\imgrowvoc{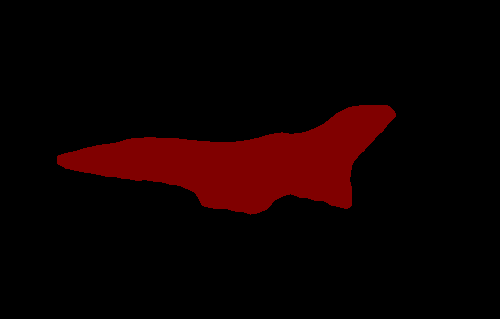}
			\imgrowvoc{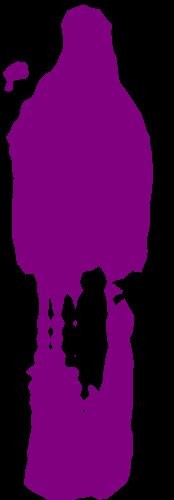}
			\imgrowvoc{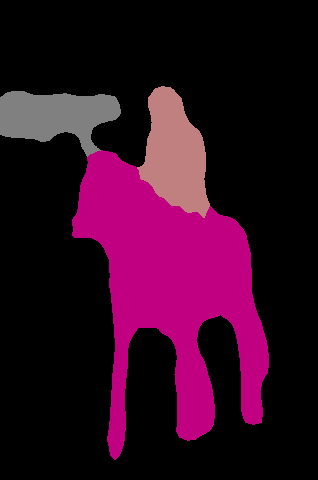}
			\imgrowvoc{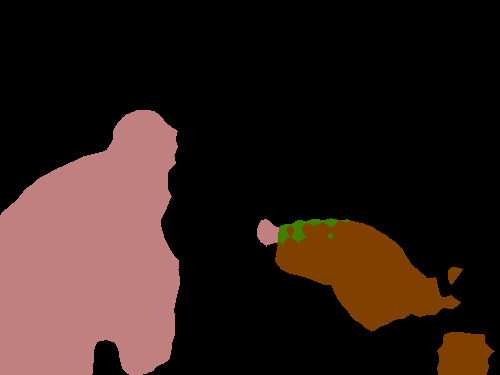}
			
			image & annotation & baseline & $+\mathcal{L}_{adv}$ & $+\mathcal{L}_{adv}+\mathcal{L}_{semi}$
		\end{tabular}
		\caption{Comparisons on the PASCAL VOC dataset using 1/2 training data.}
		\label{fig: pascal2}
	\end{figure}
	
	\newcommand{\imgrowCs}[1]{
		\hspace{-4mm}
		\includegraphics[width = 0.2\linewidth, height=0.1\linewidth]{imgs/results/img/munster_000#1_000019_leftImg8bit.png}& \hspace{-4mm}
		\includegraphics[width = 0.2\linewidth, height=0.1\linewidth]{imgs/results/gt_color/munster_000#1_000019_gtFine_color.png}& \hspace{-4mm}
		\includegraphics[width = 0.2\linewidth, height=0.1\linewidth]{imgs/results/baseline/munster_000#1_000019_leftImg8bit.png}& \hspace{-4mm}
		\includegraphics[width = 0.2\linewidth, height=0.1\linewidth]{imgs/results/Ladv/munster_000#1_000019_leftImg8bit.png}& \hspace{-4mm}
		\includegraphics[width = 0.2\linewidth, height=0.1\linewidth]{imgs/results/Ladv+Lsemi/munster_000#1_000019_leftImg8bit.png} \\
	}
	\begin{figure}[t]
		\scriptsize
		\centering
		\begin{tabular}{@{}ccccc@{}}
			\imgrowCs{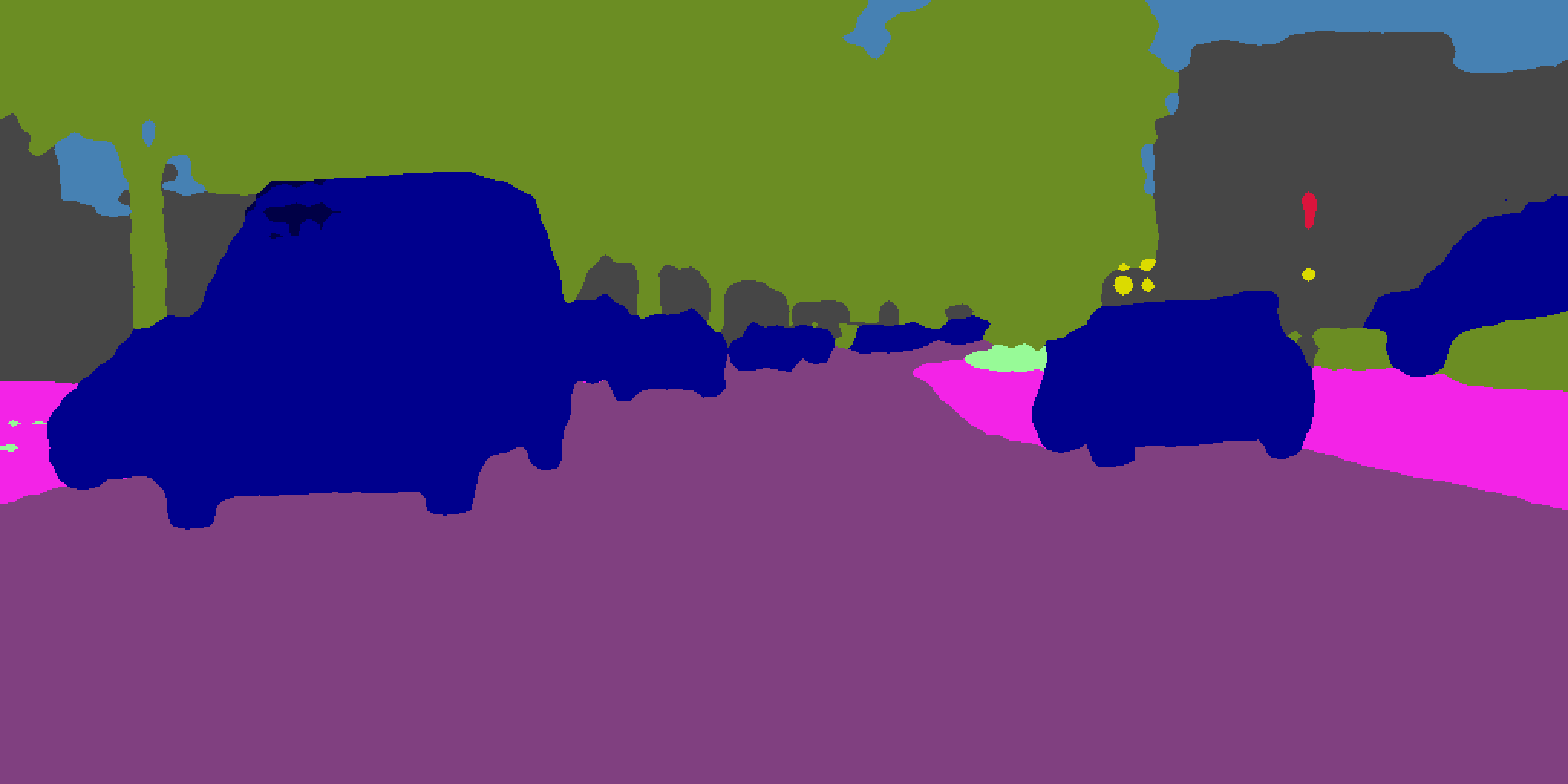}
			\imgrowCs{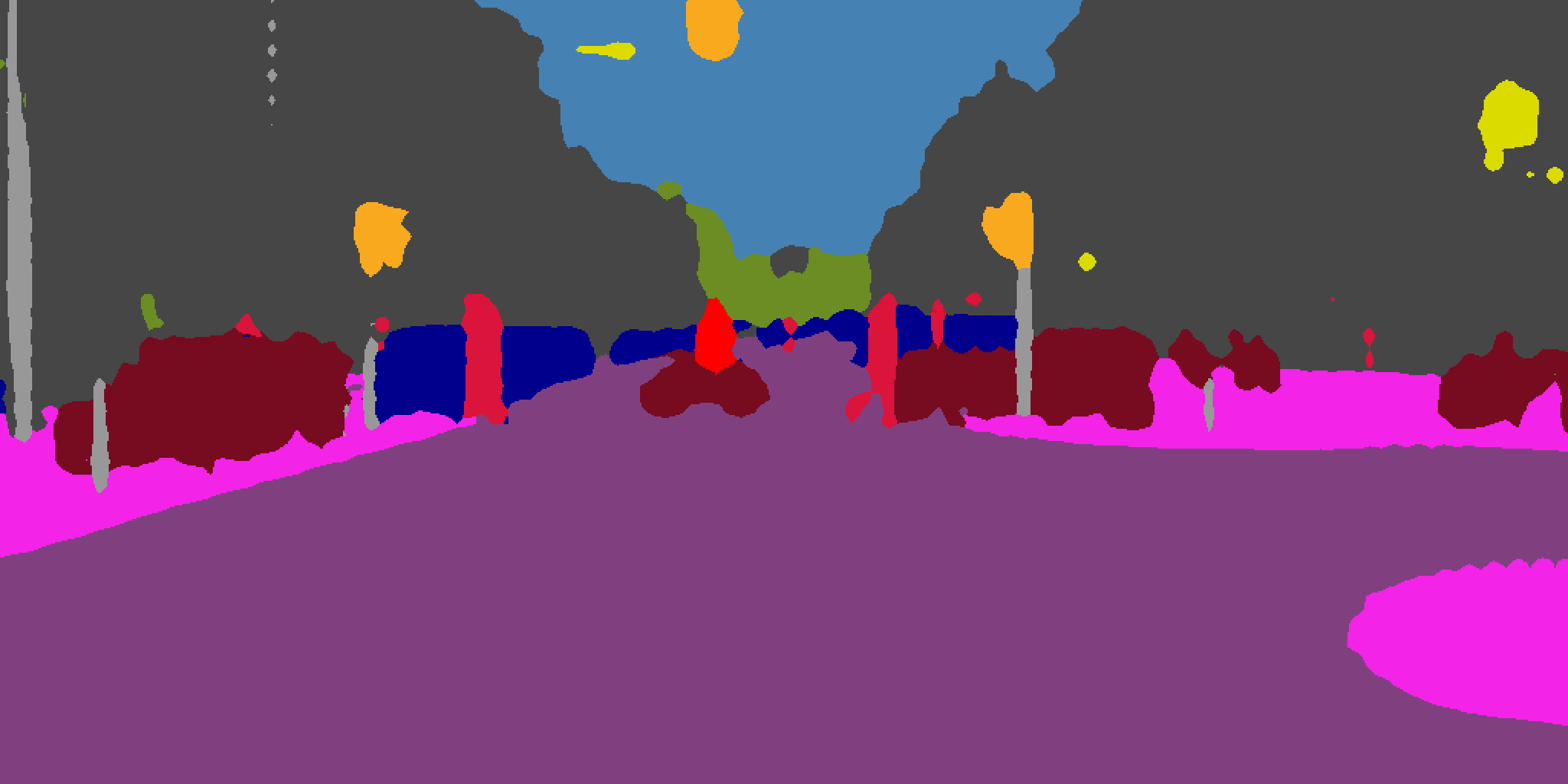}
			\imgrowCs{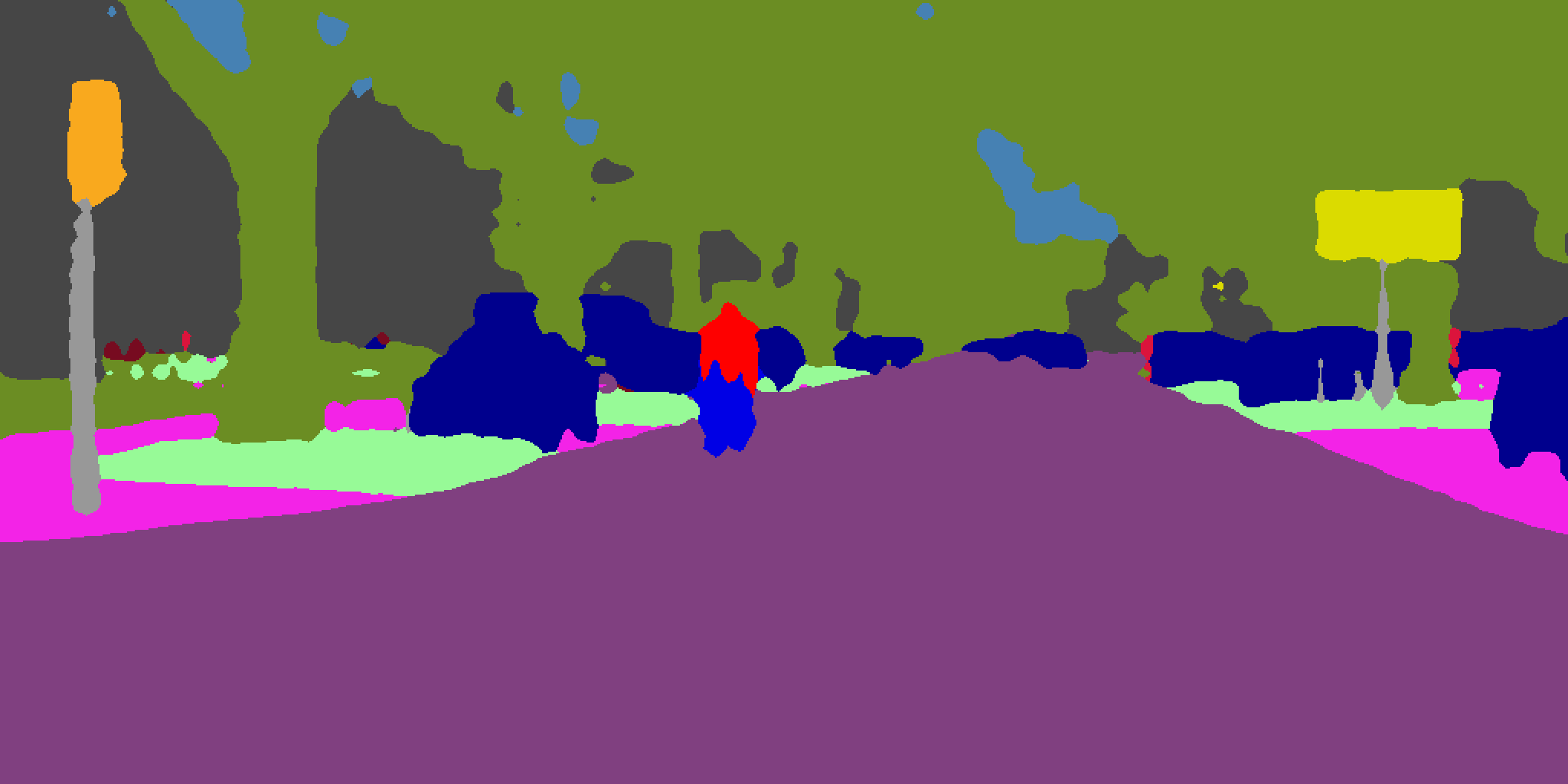}
			\imgrowCs{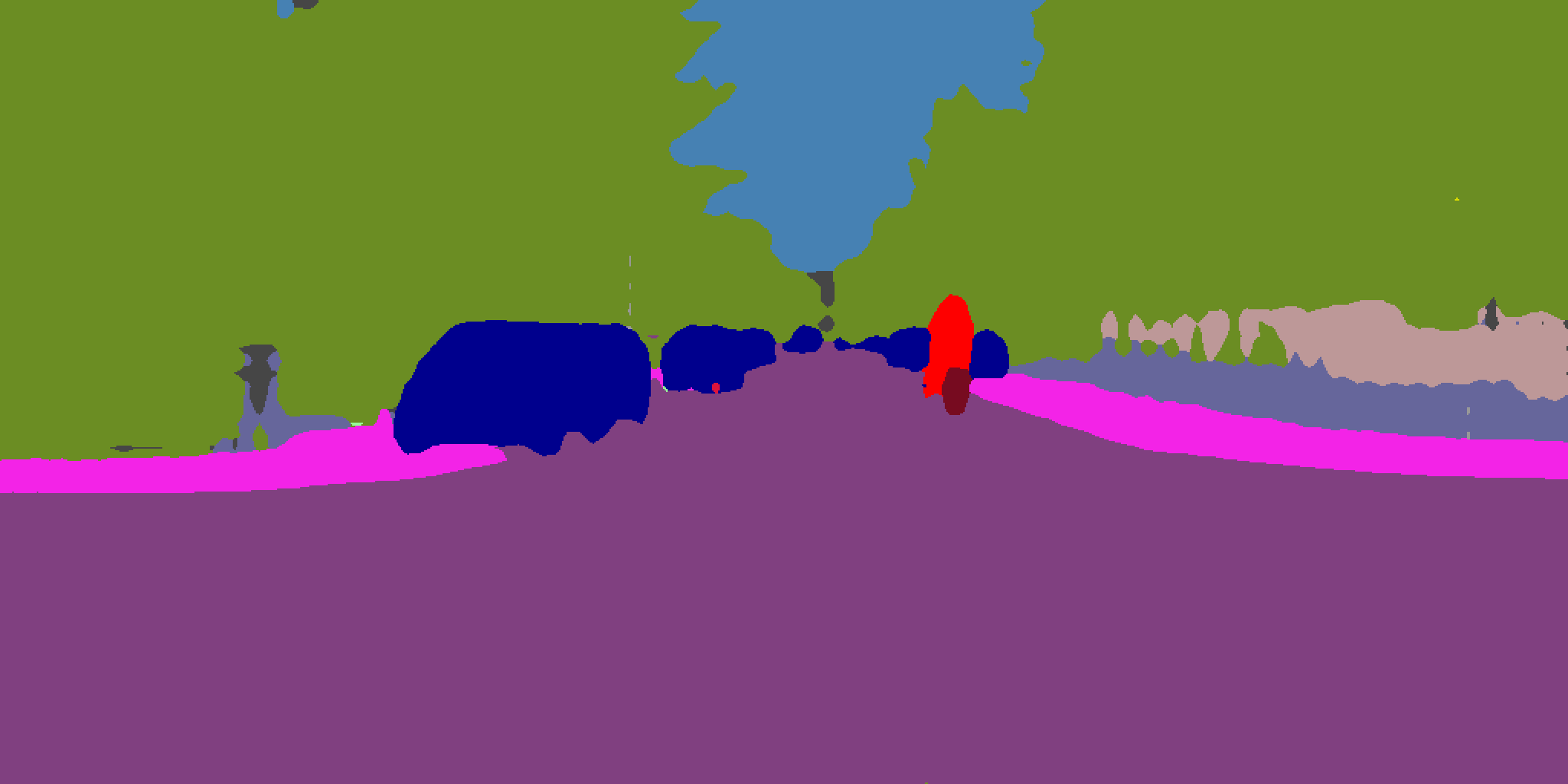}
			\imgrowCs{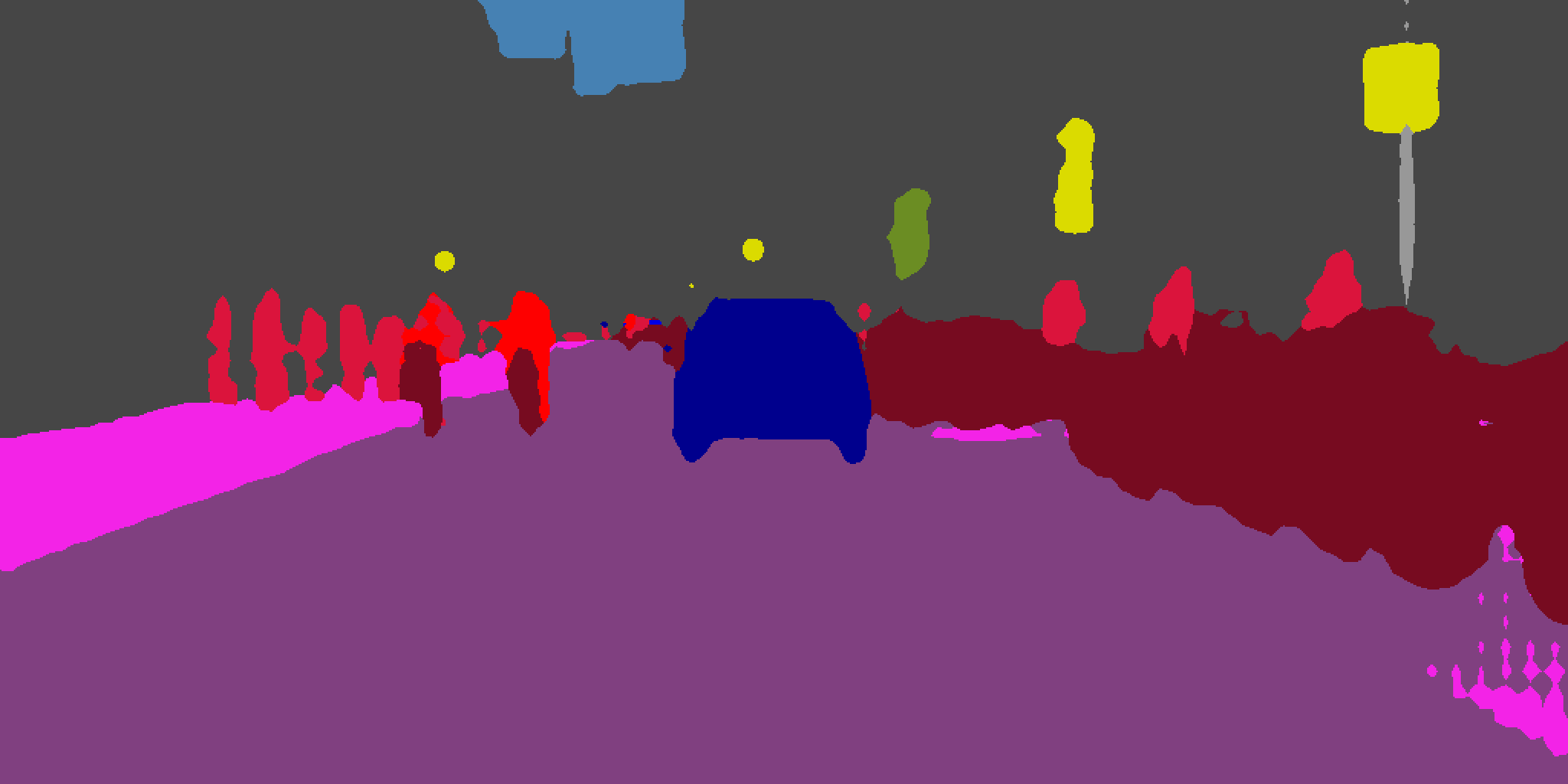}
			\imgrowCs{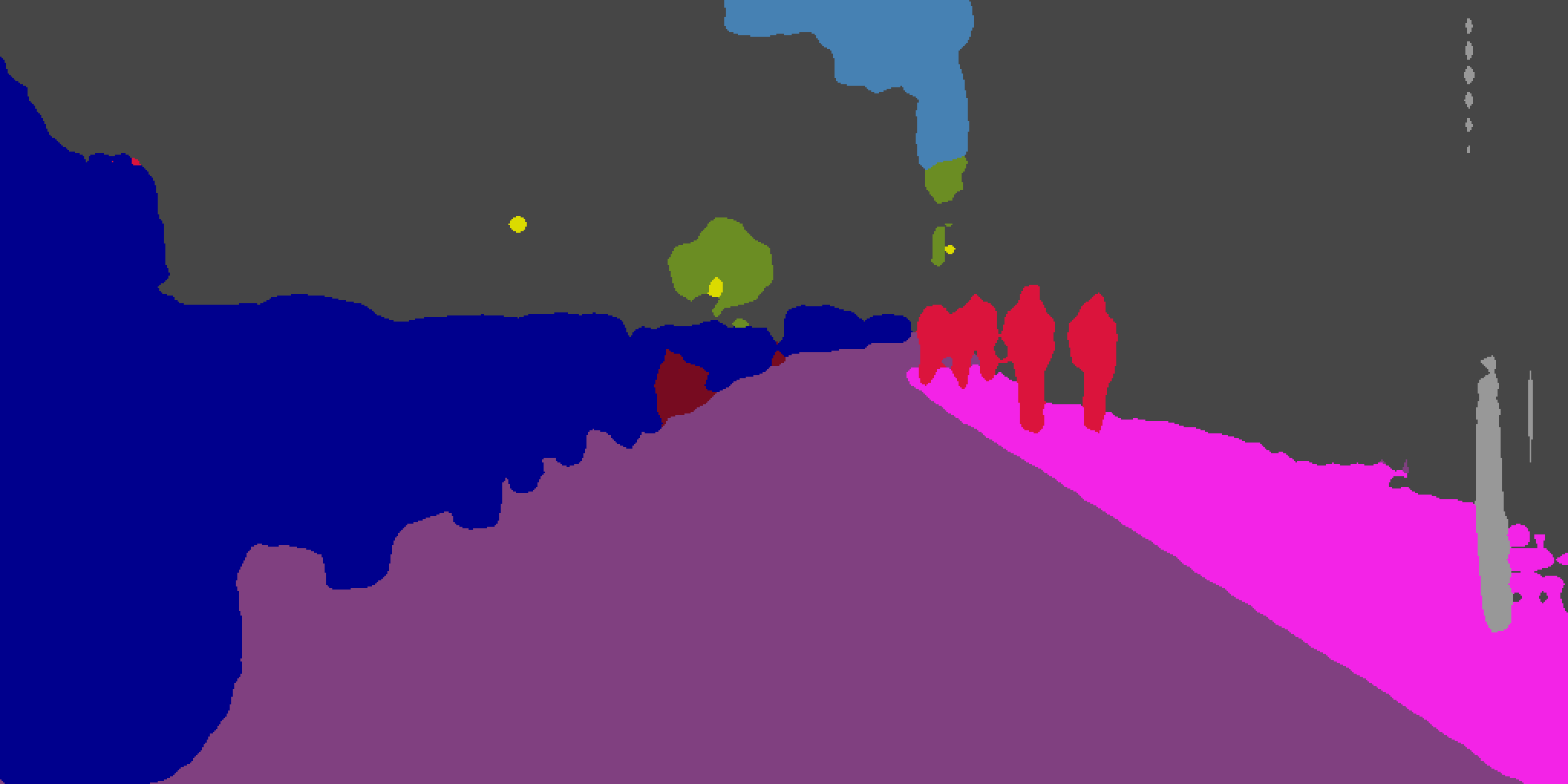}
			\imgrowCs{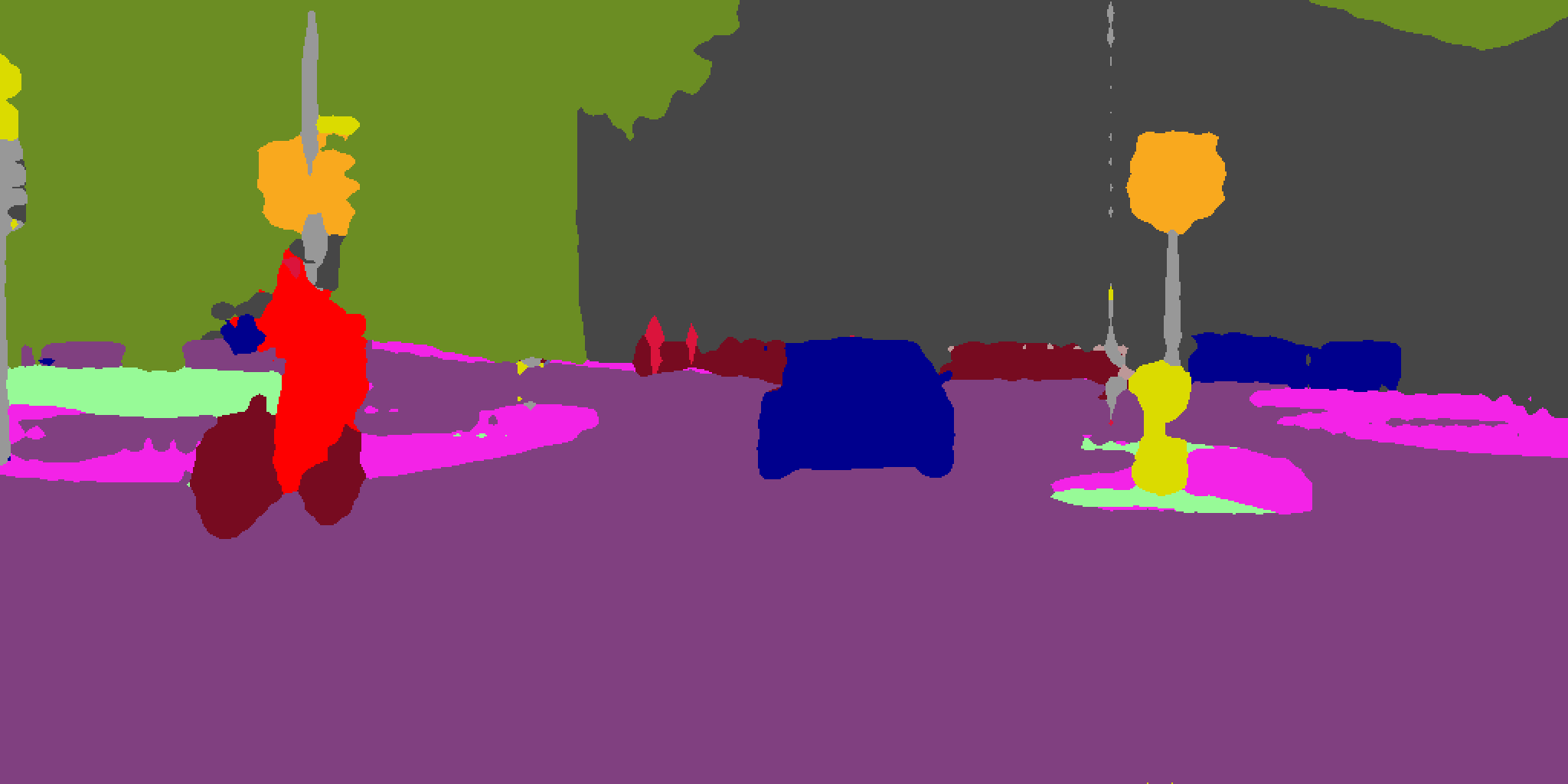}
			\imgrowCs{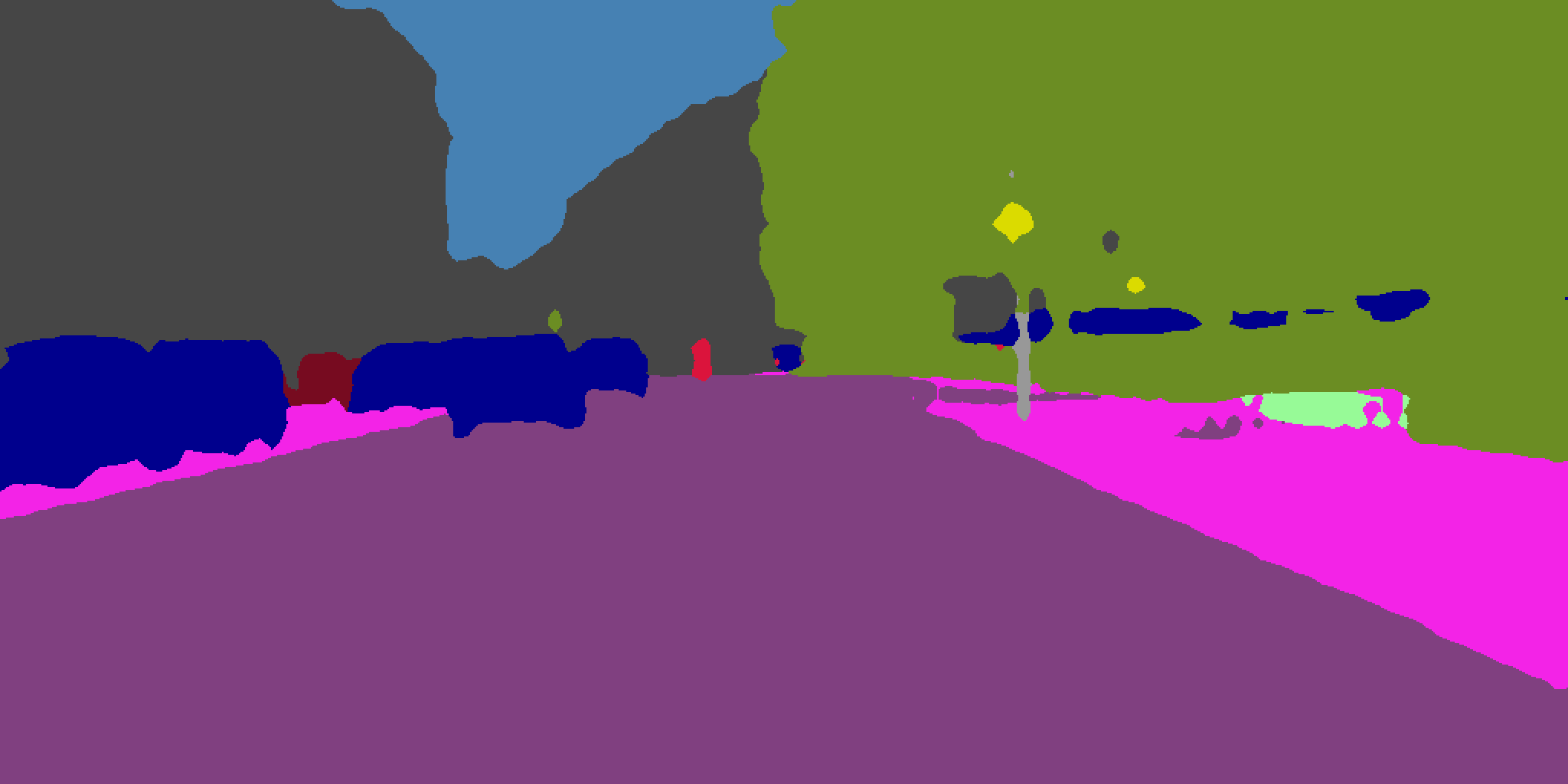}
			\imgrowCs{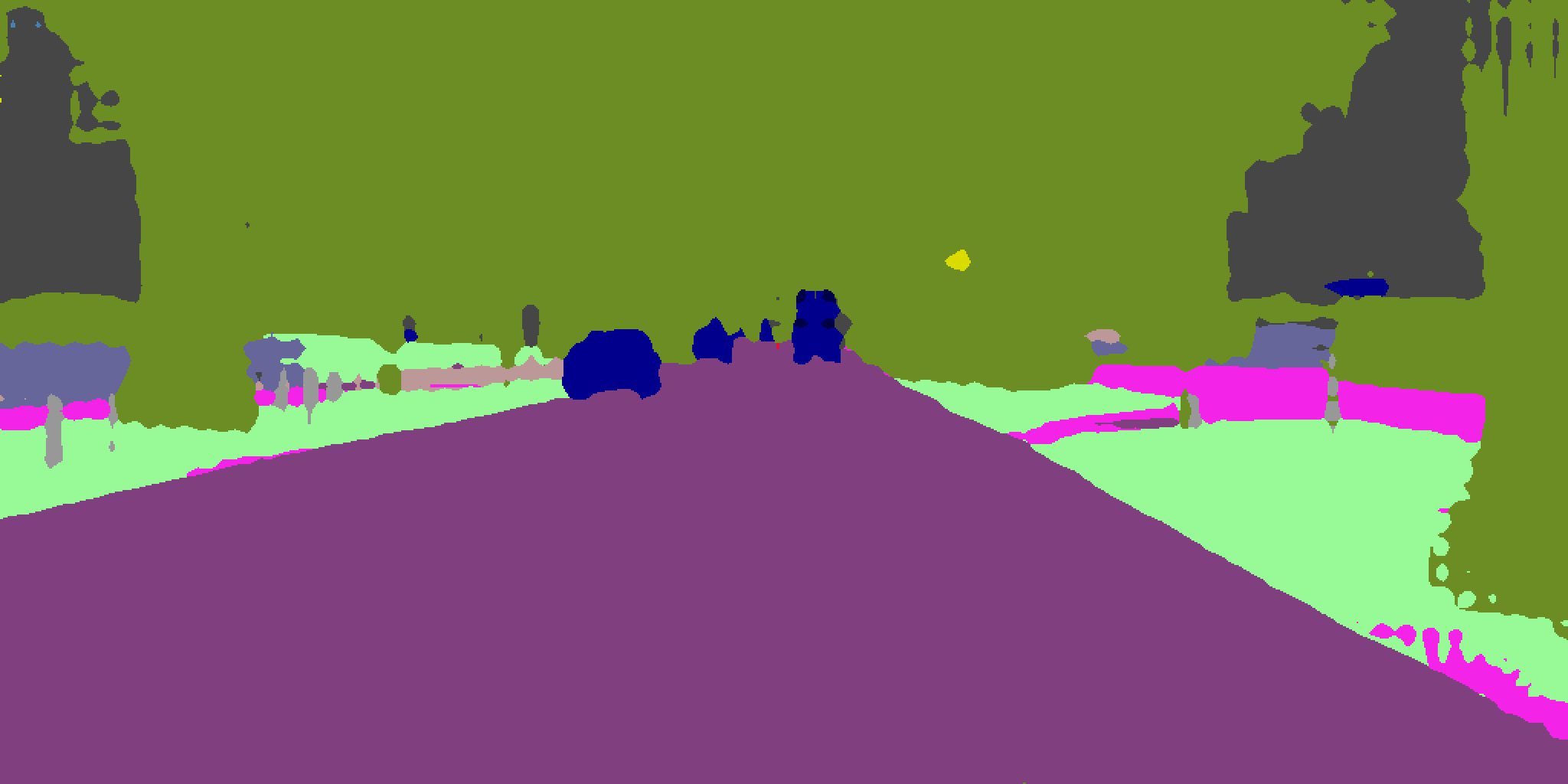}
			image & annotation & baseline & $+\mathcal{L}_{adv}$ & $+\mathcal{L}_{adv}+\mathcal{L}_{semi}$
		\end{tabular}
		\caption{Comparisons on the Cityscapes dataset using 1/2 training data.}
		\label{fig: CS1}
	\end{figure}
\end{document}